\RequirePackage[OT1]{fontenc}
\documentclass[letterpaper, 10 pt, journal, twoside]{IEEEtran}
\usepackage{ifthen}
\newcommand\arxivmode{true} 
\ifthenelse{\equal{\arxivmode}{true}}{\pdfoutput=1}{}

\usepackage{graphicx}
\DeclareGraphicsExtensions{.pdf,.png,.jpg}

\usepackage{stfloats}
\usepackage{mwe}
\usepackage{bbm}
\usepackage{amsfonts}
\usepackage{amsmath}
\usepackage{amssymb}
\usepackage{balance}
\usepackage{xcolor}
\usepackage[bookmarks=true]{hyperref}
\usepackage{url}
\usepackage{mathtools}
\usepackage{subfig}
\usepackage{siunitx}
\usepackage[ruled,linesnumbered]{algorithm2e}
\usepackage[font=small]{caption}
\usepackage{booktabs}
\usepackage{makecell}
\usepackage{pgfplots}
\usepackage{pgfplotstable}
\usepackage{fancyhdr}

\makeatletter
\let\NAT@parse\undefined
\makeatother
\usepackage[numbers,sort&compress]{natbib}
\usepackage[capitalise]{cleveref}

\newcommand{\realspace}[2][]{\mathbb{R}^{{#2}}}

\newcommand{\imagepair}{\mathcal{I}}
\newcommand{\imw}{w}
\newcommand{\imh}{h}
\newcommand{\gimage}{\mathbf{I}_g}
\newcommand{\dimage}{\mathbf{I}_d}

\newcommand{\datasetX}{\mathcal{X}}
\newcommand{\pointcloud}{\datasetX}
\newcommand{\dimension}{D}
\newcommand{\datasetY}{\mathcal{Y}}
\newcommand{\pointx}{\mathbf{x}}
\newcommand{\pointy}{\mathbf{y}}
\newcommand{\point}{\pointx}
\newcommand{\randvarX}{X}
\newcommand{\randvarY}{Y}
\newcommand{\density}[3][p]{\ensuremath{ {#1}_{{#2}} ( {#3} ) }}

\newcommand{\renyient}{H_2}
\newcommand{\csd}{D_{\text{CS}}}
\newcommand{\bigoh}{\mathcal{O}}

\newcommand{\gmm}{\mathcal{G}}
\newcommand{\weight}{\pi}
\newcommand{\mean}{\boldsymbol{\mu}}
\newcommand{\cov}{\mathbf{\Sigma}}
\newcommand{\gaussian}[4][\mathcal{N}]{\ensuremath{ {#1} ({#2} \mid {#3}, {#4})}}

\newcommand{\bmat}{\mathbf{H}}
\newcommand{\kernel}[2][K_{\bmat}]{\ensuremath{ {#1} \left( {#2} \right)}}

\newcommand{\sumoverkernel}[5][]{\ensuremath{ \sum_{{#2}}^{{#3}} \kernel{{#4} - {#5}} }}
\newcommand{\nonparamest}[4][]{\ensuremath{ \frac{1}{| {#3} |}\sum_{ {#2}=1 }^{| {#3} |} \kernel{ {#4} - {#4}_{ {#2} } } }}

\newcommand{\sogmm}[1]{SOGMM-{#1}}
\newcommand{\ndt}[1]{NDT-{#1}}
\newcommand{\om}[1]{OM-{#1}}
\newcommand{\gpom}[1]{GPOM-{#1}}
\newcommand{\fc}[1]{FC-{#1}}

\usepackage[utf8]{inputenc}
\usepackage{pgfplots}
\usepgfplotslibrary{groupplots,dateplot}
\usetikzlibrary{patterns,shapes.arrows}
\pgfplotsset{compat=newest,scaled x ticks=false}

\usepackage{tikz}
\usetikzlibrary{external,positioning}
\usepgfplotslibrary{external}
\newcommand{\blue}[1]{{\color{black}{#1}}}

\title{Probabilistic Point Cloud Modeling via Self-Organizing Gaussian Mixture Models}
\author{Kshitij Goel, Nathan Michael, and Wennie Tabib%
\thanks{Manuscript received: October, 18, 2022; Revised January, 24, 2023; Accepted February, 27, 2023.}
\thanks{This paper was recommended for publication by Javier Civera upon evaluation of the Associate Editor and Reviewers' comments.} 
\thanks{The authors are with The Robotics Institute, Carnegie Mellon University, Pittsburgh, PA 15213 USA
(email: \{\texttt{kshitij,nmichael,wtabib}\}\texttt{@cmu.edu}).}%
\thanks{Digital Object Identifier (DOI): see top of this page.}
}

\PassOptionsToPackage{hyphens}{url}\usepackage{hyperref}

\fancypagestyle{firstpagestyle}
{
  \fancyhf{}
  \fancyhead[L]{\footnotesize{%
      \begin{center}
          This paper has been accepted for publication in \emph{IEEE Robotics and Automation Letters}. \\ DOI: \href{https://doi.org/10.1109/LRA.2023.3256923}{10.1109/LRA.2023.3256923} \\
        \end{center}
    }}
  \fancyfoot[L]{\footnotesize{%
      \vspace{-0.5cm}
        \textcopyright 2023 IEEE. Personal use of this material is
        permitted. Permission from IEEE must be obtained for all other uses,
        in any current or future media, including reprinting/republishing this
        material for advertising or promotional purposes, creating new
        collective works, for resale or redistribution to servers or lists, or
        reuse of any copyrighted component of this work in other works.
  }}

}

\begin{document}
\ifthenelse{\equal{\arxivmode}{true}}
{
\twocolumn[{%
\begin{@twocolumnfalse}
\maketitle
\begin{minipage}{\textwidth}
  \centering
  \includegraphics[width=0.33\textwidth,height=4cm,trim={300 125 300 125},clip]{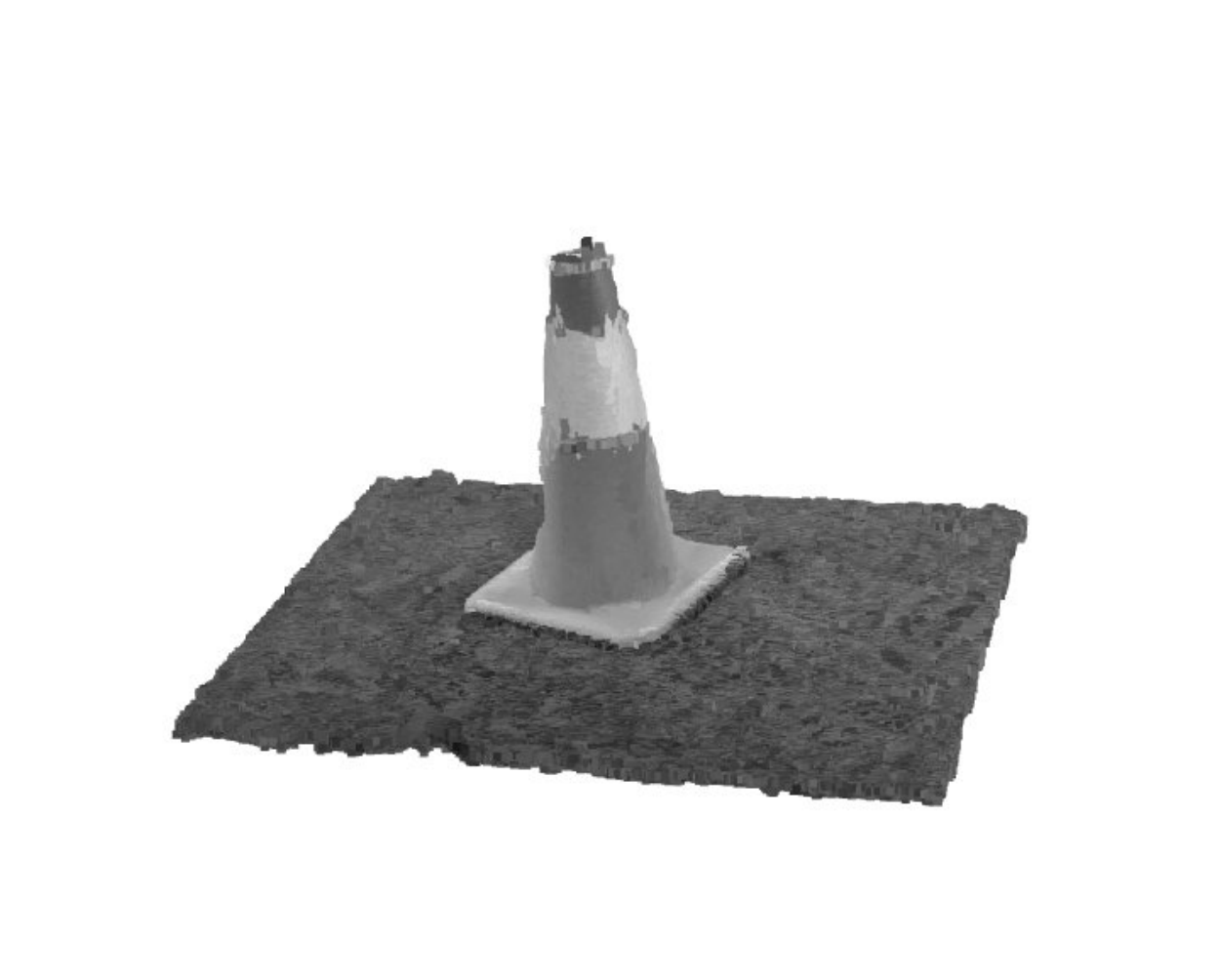}%
  \includegraphics[width=0.33\textwidth,height=4cm,trim={100 50 100 50},clip]{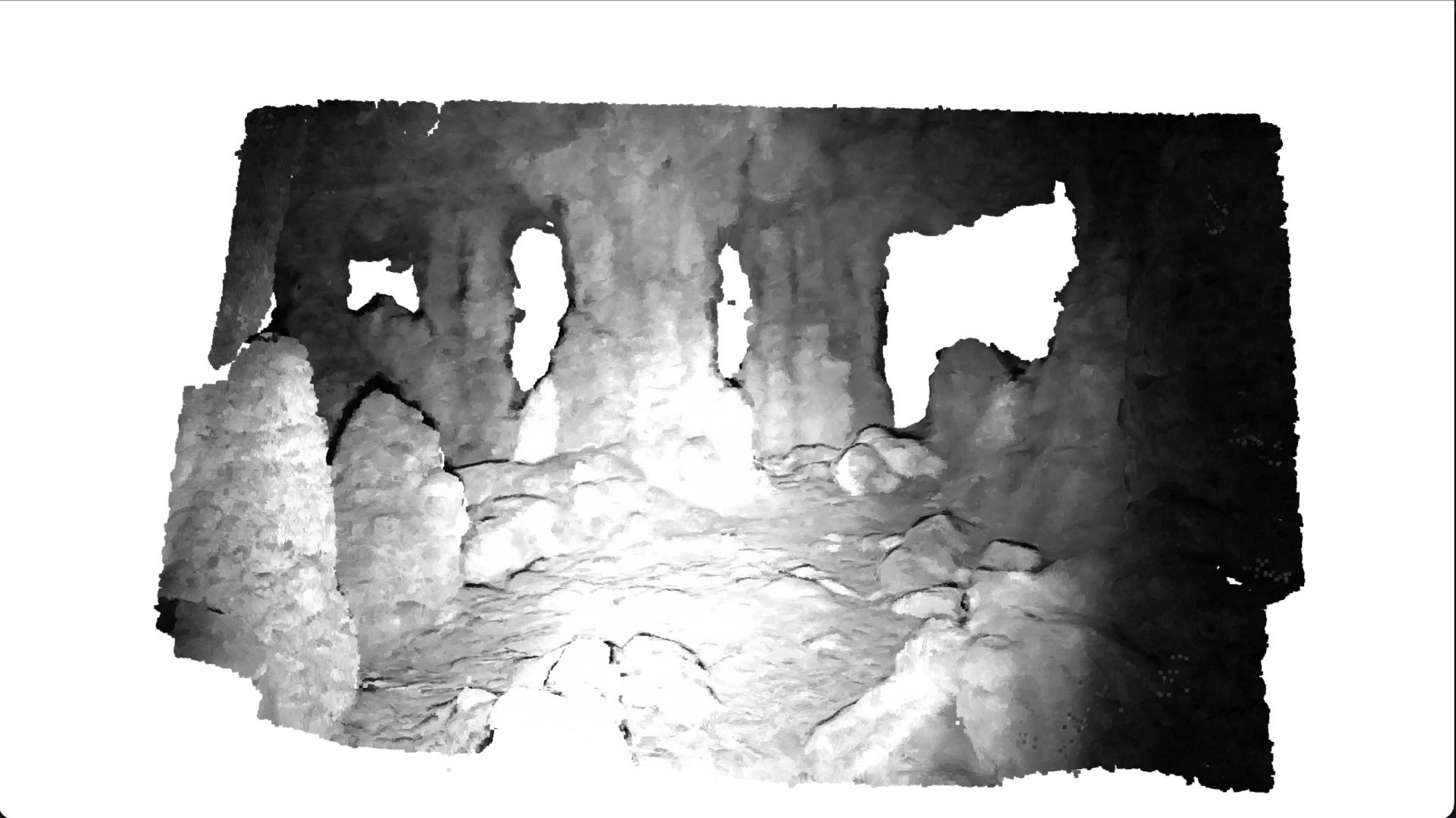}%
  \includegraphics[width=0.33\textwidth,height=4cm,trim={0 0 0 0},clip]{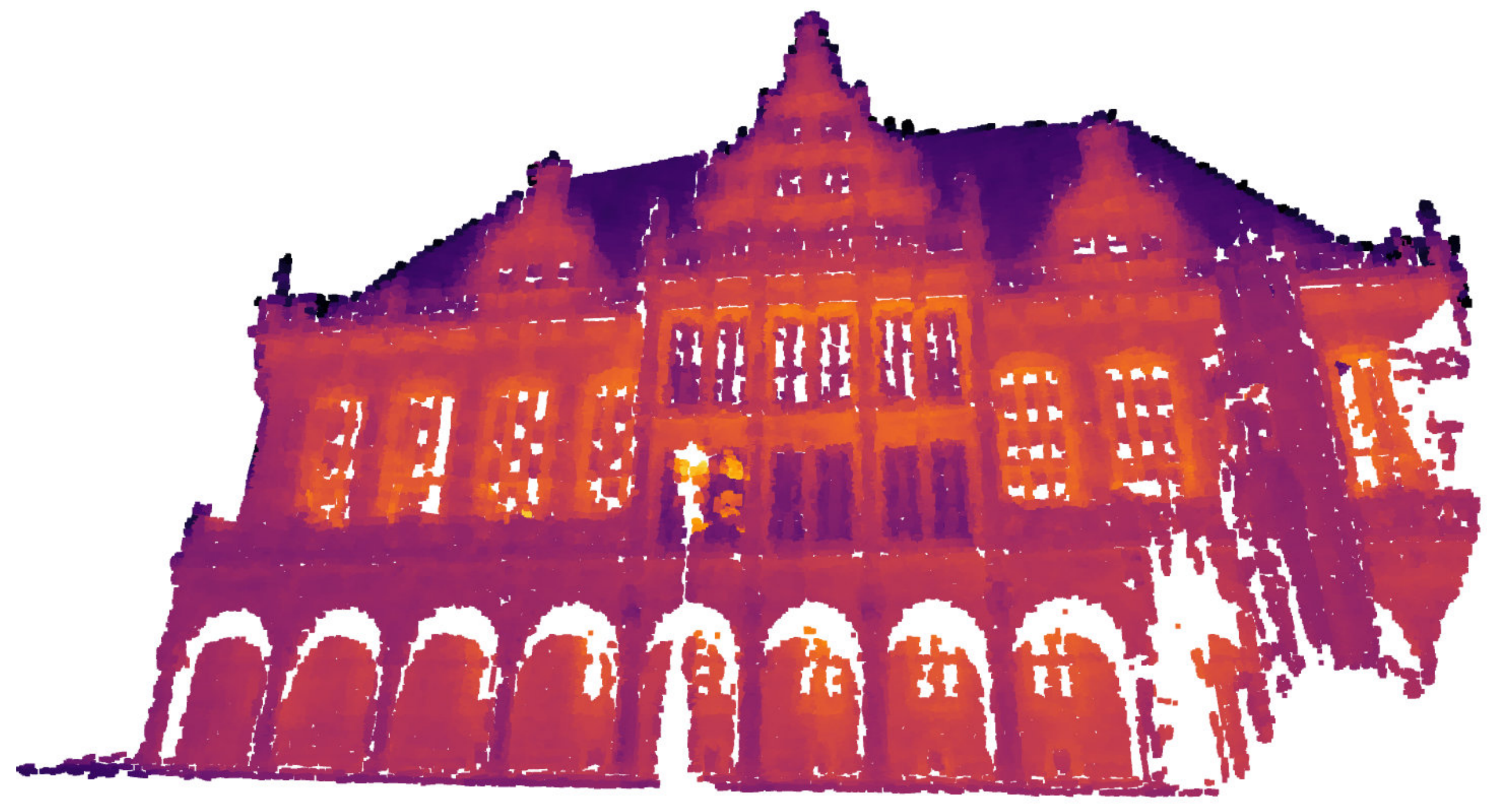}%
  \captionof{figure}{\label{fig:glory-shot} The methodology proposed in this
  work enables multi-modal reconstruction at varying scales.
  Without modifying parameters, the methodology models depth and grayscale data
  of small objects ($\SI{1}{\meter} \times \SI{1}{\meter}$ safety cone in the left image) and complex
  environments ($\SI{10}{\meter} \times \SI{5}{\meter}$ cave in the center image) while also
  modeling depth and thermal data of (right) large-scale buildings
  ($\SI{42}{\meter} \times \SI{28}{\meter}$)~\citep{3dtk_robotic_scan}.
  A supplementary video may be found at \url{https://youtu.be/v0DfhK1lyno}.}
\end{minipage}
\end{@twocolumnfalse}
}]
{
  \renewcommand{\thefootnote}%
  {\fnsymbol{footnote}}
  \footnotetext[1]{The authors are with The Robotics Institute, Carnegie Mellon University, Pittsburgh, PA 15213 USA
 (email: \{\texttt{kshitij,nmichael,wtabib}\}\texttt{@cmu.edu}).}
}
}%
{
\maketitle
\begin{figure*}
\begin{minipage}{\textwidth}
  \centering
  \includegraphics[width=0.33\textwidth,height=4cm,trim={300 125 300 125},clip]{images/cone_recon.eps}%
  \includegraphics[width=0.33\textwidth,height=4cm,trim={100 50 100 50},clip]{images/cave_recon.eps}%
  \includegraphics[width=0.33\textwidth,height=4cm,trim={0 0 0 0},clip]{images/3dtk_glory.eps}%
  \captionof{figure}{\label{fig:glory-shot} The methodology proposed in this
  work enables multi-modal reconstruction at varying scales.
  Without modifying parameters, the methodology models depth and grayscale data
  of small objects ($\SI{1}{\meter} \times \SI{1}{\meter}$ safety cone in the left image) and complex
  environments ($\SI{10}{\meter} \times \SI{5}{\meter}$ cave in the center image) while also
  modeling depth and thermal data of (right) large-scale buildings
  ($\SI{42}{\meter} \times \SI{28}{\meter}$)~\citep{3dtk_robotic_scan}.
  A supplementary video may be found at \url{https://youtu.be/v0DfhK1lyno}.}
\end{minipage}
\end{figure*}
}

\ifthenelse{\equal{\arxivmode}{true}}
{\thispagestyle{firstpagestyle}}%
{\markboth{IEEE Robotics and Automation Letters. Preprint Version. Accepted February, 2023}
{Goel \MakeLowercase{\textit{et al.}}: Probabilistic Point Cloud Modeling via Self-Organizing Gaussian Mixture Models}}

\begin{abstract}
  This letter presents a continuous probabilistic modeling methodology for
  spatial point cloud data using finite Gaussian Mixture Models (GMMs) where the
  number of components are adapted based on the scene complexity. Few
  hierarchical and adaptive methods have been proposed to address the
  challenge of balancing model fidelity with size. Instead, state-of-the-art
  mapping approaches require tuning
  parameters for specific use cases, but do not generalize across diverse
  environments.
  To address this gap, we utilize a self-organizing principle from
  information-theoretic learning to automatically adapt the complexity of the
  GMM model based on the relevant information in the sensor data. The
  approach is evaluated against existing point cloud modeling techniques on
  real-world data with varying degrees of scene complexity.
\end{abstract}

\begin{IEEEkeywords}
  Mapping, RGB-D Perception, Field Robots
\end{IEEEkeywords}

\section{INTRODUCTION}\label{sec:intro}
\IEEEPARstart{D}{ense} point cloud data are used in
physical simulations~\citep{Ummenhofer2020Lagrangian}, computer
graphics~\citep{vedaldi_neural_2020}, and robotic
perception~\citep{tabib_autonomous_2021}.  For robotic perception applications,
in particular, three-dimensional (3D) perception algorithms do not operate
directly on raw point cloud data; instead, they subsample, discretize, or create
an intermediate representation~\cite{eckart_compact_2017}. Gaussian mixture models (GMMs) have been
proposed as a generative model to compactly parameterize raw point cloud
data~\citep{eckart_accelerated_2016,srivastava_efficient_2019-1,omeadhra_variable_2019}
and enable high-resolution transmission of perceptual information to other
robotic systems in communication restricted
environments~\citep{goel_rapid_2021,corah_communication-efficient_2019}.
These models have been leveraged to represent
occupancy~\cite{omeadhra_variable_2019}, estimate pose~\cite{tabib2018manifold}, and perform incremental
mapping~\cite{srivastava_efficient_2019-1,dhawale_efficient_2020}, which are
fundamental subsystems for higher-level robot autonomy.
However, these prior works require model complexity selection
before operation and do not adapt based on three-dimensional structure.
We address this gap by proposing a GMM-based
point cloud modeling technique which adapts the number of components
according to the scene complexity~(\cref{fig:glory-shot}).

The fundamental challenge is to estimate the number of components
required to model the point cloud to obtain sufficient fidelity.
~\citet{mclachlan_number_2014} provide an overview of approaches to
estimate the number of components in a GMM. Among the
commonly used criteria are the Akaike's information criterion
(AIC)~\citep{akaike_new_1974}, Bayesian information criterion
(BIC)~\citep{schwarz_estimating_1978}, and Model description length
(MDL)~\citep{hansen_model_2001}, which strike a balance
between the fit over the underlying dataset and model complexity. For
example, to use AIC or BIC scores for model selection one must plot the
scores over many candidate mixtures with increasing numbers of components to
detect an approximate minima, which is prohibitive for online robotics applications
with finite computational power and timing constraints. Variational methods have been
created to estimate \emph{a posteriori} distributions over the parameters of a GMM;
however, these methods also require specifying a prior distribution over model
parameters~\citep{attias_variational_1999,blei_variational_2006}.

\textbf{Contributions.} In contrast to these methods, this work
proposes a perceptual modeling methodology that reduces the
redundancy in a dataset and extracts the \emph{relevant} information
via a self-organizing principle called the Principle of Relevant
Information (PRI). We call this approach the \emph{Self-Organizing
Gaussian Mixture Modeling} (SOGMM) method as it adapts the number of
components of a GMM via self-organization. The rest of the paper is
structured as follows.~\Cref{sec:related-work} reviews related work in
point cloud modeling for robotic perception.~\Cref{sec:approach}
presents the details of the SOGMM method and~\cref{sec:evaluation}
provides evaluation on real-world point cloud data. We conclude
this letter and discuss future work in~\cref{sec:conclusion}.

\section{RELATED WORK}\label{sec:related-work}
In this section, we focus on reviewing methods that enable
probabilistic modeling of multi-modal point cloud data for high-fidelity
reconstruction.

\textbf{Discrete Methods.}~Robotics applications commonly leverage
probabilistic environment representations consisting of fixed-size
volumetric pixels (or voxels in 3D) and represent occupancy as
independent, discrete random
variables~\cite{elfes_using_1989}. The discrete probability distribution is
updated using the inverse sensor model and Bayes' rule.  The limitation
of this representation is it requires specifying parameters such as
voxel size and occupancy clamping thresholds, which are dependent on
environmental conditions. Although the compute usage for Bayesian
updates is low, voxel grids require significant memory to scale
in spatially-extensive environments. Multi-modal extensions for 3D voxel
grids suffer from the same limitations~\citep{tabib_efficient_2016}.

To reduce the number of voxels,~\citet{magnusson_scan_2007} propose the
Normal Distributions Transform mapping framework (NDTMap) that
utilizes the same volumetric factorization but places a Gaussian distribution in
each occupied voxel. The Voxblox method by~\citet{oleynikova_voxblox_2017} uses
truncated signed distance functions (TSDFs) over regular 3D voxel grids for
point cloud modeling as opposed to an occupancy-based formulation. Both of these
frameworks model the point cloud with less memory than regular 3D voxel grids,
but still require specifying a map resolution in advance.

\citet{hornung_octomap_2013} present OctoMap to address the
limitation regarding pre-specified voxel resolution via the hierarchy
of an octree. The limitations are that pre-specifying
a minimum leaf size, which determines the maximum size of
the map, and occupancy thresholds are still required. Recent advances over OctoMap
demonstrate improvements in compression and
fidelity~\citep{duberg_ufomap_2020,funk_multi-resolution_2021};
however, the limitations regarding pre-specified parameters are
unaddressed.

In contrast to these discrete methods, we present a continuous
probabilistic multi-modal modeling methodology that adapts the model
complexity according to the scene variation.

\textbf{Continuous Methods.} Continuous non-parametric point cloud modeling
methods like Gaussian Process (GP)~\citep{ocallaghan_gaussian_2012} and
Hilbert maps~\citep{ramos_hilbert_2016} increase
representational power at the expense of high computational cost.
Closest to our work,~\citet{zobeidi_dense_2022} develop an incremental
metric-semantic GP mapping approach that represents map uncertainty
and outperforms deep neural network-based approaches. However, the need for
offline training and specification of model hyperparameters precludes the
use of this representation for online adaptive compression.
~\citet{yan_online_2021} present a highly-parallelized
adaptive scene representation that utilizes a Dirichlet Process (DP) mixture
model as a point cloud model. However, an upper limit on the number of components
allowed per processor must be pre-specified, which limits the adaptability
of the resulting model.

Neural Implicit Representations (NIRs) have been proposed as continuous implicit
models for radiance and occupancy information. Most works in this area
build on Neural Radiance Fields (NeRFs) proposed
by~\citet{mildenhall_nerf_2020}. While NeRFs produce
high-resolution, realistic models, they suffer from one major limitation due to
the fixed architecture of the underlying neural network, which does not allow
adaptivity in the representation of the scene.

Generative probabilistic models that use a finite mixture of probability
distributions aim to represent the environment through an adaptive and
parametric mathematical model, and thus provide a potential solution to the
limitation in adaptivity of NeRFs. Recently, Gaussian mixture models
(GMMs) have been utilized for adaptive point cloud
modeling~\citep{eckart_accelerated_2016,navarrete_compression_2018,
srivastava_efficient_2019-1,omeadhra_variable_2019,dong_mr-gmmapping_2022}.
\citet{eckart_accelerated_2016} use a top-down hierarchy of 3D GMMs to represent
3D point cloud data at different levels of
detail.~\citet{srivastava_efficient_2019-1} use a bottom-up hierarchy of 4D GMMs
to model the surface point clouds and the points along the sensor
beams.~\citet{navarrete_compression_2018} modify the FastGMM framework
by~\citet{greggio_fast_2012} and use it for compressing locally planar point
clouds. The FastGMM approach creates a GMM via a deterministic splitting
criterion in a coarse-to-fine manner.~\citet{dong_mr-gmmapping_2022} use a
similar idea for adaptive model selection in their GMM mapping approach.
The Gaussian components belonging to the same plane are merged using their mean
vectors and principal directions to obtain a coarser model. Common to all these works is a
requirement to specify the levels of detail; either
directly (e.g., number of layers in hierarchy of GMMs)~\citep{eckart_accelerated_2016} or
indirectly (e.g., through convergence, splitting, or merging criteria)
~\citep{srivastava_efficient_2019-1,navarrete_compression_2018,dong_mr-gmmapping_2022}.
Specifying these parameters \emph{a priori} can be challenging for real-world
robotic applications.

In contrast to these methods, the proposed approach does not require processing
a hierarchy of GMMs to achieve online adaptation with respect to scene complexity.
Instead, we perform model selection for GMMs through information-theoretic learning
as described in the next section.


\section{APPROACH}\label{sec:approach}
\begin{figure}
  \ifthenelse{\equal{\arxivmode}{true}}%
  {\includegraphics[width=\columnwidth]{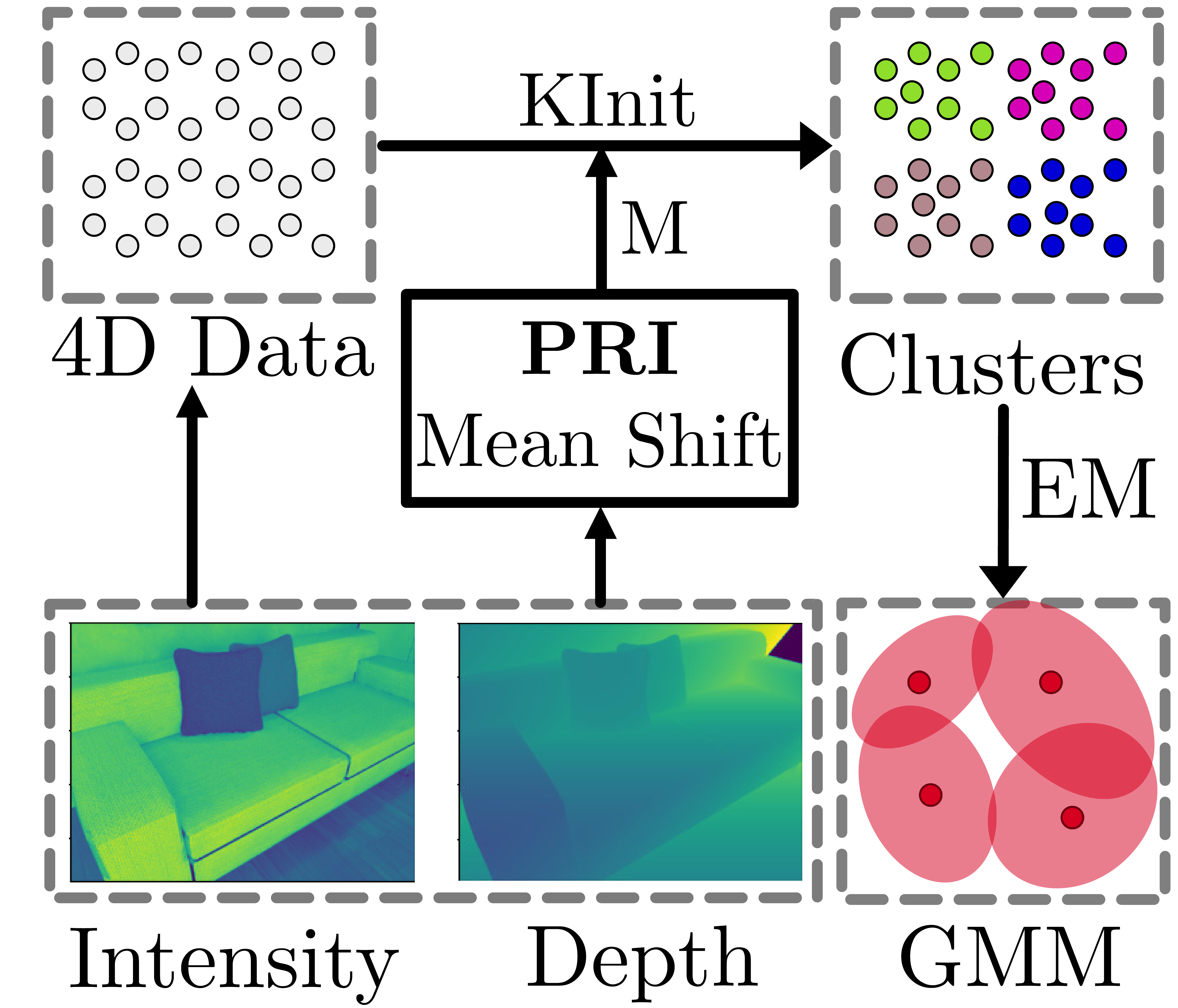}}%
  {\includegraphics[width=\columnwidth]{images/overview.eps}}
  \caption{\label{fig:overview}Overview of the proposed point cloud modeling method. The registered
    intensity-depth image pair is used by the PRI component to determine the number
    of modes $M$. Steps (1a)--(1c) of the K-Means++ algorithm~\citep{arthur_k-means_2007} perform a hard-partitioning of the 4D data into $M$
    clusters (KInit). Finally, the EM algorithm does a soft-partitioning of the 4D data using the KInit
    output to create a $M$-component finite GMM.
    The proposed system encodes the 4D point cloud data into a finite GMM without
    requiring the specification of number of mixtures $M$ for every image pair.}
  \vspace{-0.5cm}
\end{figure}

\textbf{Notation.} In this section, small letters are used for scalars and
univariate random variables (e.g., $\imw$, $\imh$, $x$, $g$),
capital letters for multivariate random variables (e.g., $\randvarX$, $\randvarY$),
bolded small letters for vectors (e.g., $\pointx$, $\pointy$),
bolded capital letters for matrices (e.g., $\mathbf{X}$, $\mathbf{Y}$), and
calligraphic letters for sets (e.g., $\imagepair$, $\datasetX$, $\datasetY$).
The probability density of a continuous multivariate random variable
$\randvarX$ is written as $\density{\randvarX}{\pointx}$, where $\pointx$
is a value in the sample space of $\randvarX$.

\textbf{Overview.} An overview of the SOGMM system is shown in~\cref{fig:overview}. We assume that
a registered pair of depth, $\dimage \in \realspace{\imh \times \imw}$, and grayscale,
$\gimage \in \realspace{\imh \times \imw}$, images $\imagepair \coloneqq \{\dimage, \gimage\}$
of dimension $\imh \times \imw$ is provided for point cloud modeling.
Note that the grayscale modality can also be thermal data (e.g.,
the building in~\cref{fig:glory-shot}) scaled to a range $[0, 1]$.
The image frame points may
be projected to three dimension using the intrinsic camera matrix and associated with
depth to obtain a pointcloud $\pointcloud$ with $hw$ points, where
each point $\point_i \in \pointcloud$ is four-dimensional and consists of 3D coordinates
$(x, y, z)$ augmented with a grayscale value $g$. Like prior works in GMM-based
point cloud modeling~\citep{eckart_accelerated_2016,srivastava2017efficient,tabib_autonomous_2021},
we assume that the points in $\datasetX$ are independently
and identically distributed (i.i.d.) samples from an underlying continuous
random variable $\randvarX$. The goal of the proposed approach is to model the
joint probability distribution $p(x, y, z, g)$ as a GMM, $\gmm(\imagepair)$.
Thus, the probability density for $\gmm(\imagepair)$ is written as:
\begin{align}
  \gmm(\imagepair) \equiv \density{\randvarX}{\pointx} = \sum_{m=1}^M \weight_m \gaussian{\pointx}{\mean_m}{\cov_m},
  \label{eq:model}
\end{align}
where, $\weight_m \in \realspace{}$, $\mean_m \in \realspace{4}$, and
$\cov_m \in \realspace{4 \times 4}$ are the weight, mean, and covariance
associated with the $m^{\text{th}}$ multivariate Gaussian distribution component
of the mixture. In contrast to existing techniques where the number of mixtures
is specified \emph{a priori}~\citep{tabib_real-time_2019,corah_communication-efficient_2019} or
utilize a hierarchy of GMMs to estimate the number of
components~\citep{eckart_accelerated_2016,srivastava2017efficient},
the contribution of this work is a self-organizing approach that learns the
number of mixtures from the underlying sensor data, $\imagepair$, via
information-theoretic techniques.  This learned value is used for
generative modeling for dataset $\datasetX$ via the Expectation-Maximization
(EM) algorithm~\citep{bilmes1998gentle}, and utilizes
steps (1a)--(1c) of K-Means++~\citep{arthur_k-means_2007} for initialization. The
output of EM is the model $\gmm(\imagepair)$ from~\cref{eq:model}. Since this model
is a joint distribution over the 3D spatial coordinates and the grayscale data,
we can use the conditional distribution $p(g \mid \begin{bmatrix} x & y & z\end{bmatrix}^{\top})$
to regress the grayscale image from the model for a given spatial point cloud
and pinhole camera model~\citep{sung_gaussian_2004,srivastava2017efficient}. The
remainder of this section describes how to estimate the number of mixture components
$M$.

\textbf{Principle of Relevant Information.} The
Principle of Relevant Information (PRI)~\citep{principe_information_2010} is an approach
that extracts the relevant statistics from the dataset to learn a compressed representation
of size $M$, equal to the number of locally-dense regions in the environment. The
intuition behind PRI is to extract these relevant statistics by simultaneously
minimizing the redundancy and distortion between the original and compressed datasets.
Formally, let us consider a dataset, $\datasetY$, in which the points are assumed to be $\dimension$-dimensional
i.i.d. samples from a continuous multivariate random variable $\randvarY$
with the associated probability density function $\density{\randvarY}{\pointy}$. To create a
compressed dataset $\datasetY_r$, with random variable $\randvarY_r$ and density $\density{\randvarY_r}{\pointy}$,
the PRI is an information-theoretic optimization problem with the objective function:
\begin{align}
  J(\randvarY_r) = \min_{\randvarY_r} \renyient (\randvarY_r) + \csd (\randvarY_r, \randvarY),
  \label{eq:pri}
\end{align}
where, $\renyient (\randvarY_r)$ is the Renyi's quadratic entropy (RQE) of
dataset $\datasetY_r$ calculated using the density $\density{\randvarY_r}{\pointy}$ and
$\csd (\randvarY_r, \randvarY)$ is the Cauchy-Schwarz divergence (CSD)
between datasets $\datasetY_r$ and $\datasetY$ calculated using the densities $\density{\randvarY_r}{\pointy}$
and $\density{\randvarY}{\pointy}$. Minimizing RQE
ensures less redundancy in the compressed dataset and minimizing CSD reduces the
error induced due to compression.~\citet{principe_information_2010} proves that when
RQE and CSD have an equal contribution to the objective function in~\cref{eq:pri}, the compressed
dataset $\datasetY_r$ contains the modes of the original dataset $\datasetY$.

\textbf{Gaussian Mean Shift.} \citet{rao_mean_2009} show that the Gaussian Mean Shift
(GMS) algorithm, as proposed by~\citet{cheng_mean_1995}, is an iterative scheme
for an approximate solution to the optimization problem in~\cref{eq:pri}.
The GMS algorithm uses a nonparametric estimate of the density $\density{\randvarY}{\pointy}$,
\begin{align}
  \density{\randvarY}{\pointy} = \nonparamest{i}{\datasetY}{\pointy},
\end{align}
where, $\bmat$ is a symmetric positive definite $\dimension \times \dimension$
matrix and $\kernel{\pointy - \pointy_i}$ is a multivariate symmetric Gaussian kernel.
This matrix is usually chosen proportional to the identity matrix $\bmat = \sigma^2 \mathbf{I}_{D \times D}$,
where $\sigma$ is called the \textit{bandwidth} parameter~\citep{comaniciu_mean_2002}.
Under this assumption the kernel $\kernel{\pointy - \pointy_i}$ simplifies to:
\begin{align}
  \kernel{\pointy - \pointy_i} = \frac{1}{(2 \pi)^{\dimension / 2} \sigma} \exp \left( -\frac{1}{2 \sigma^2} \| \pointy - \pointy_i \|^2 \right).
  \label{eq:gaussiankernel}
\end{align}
The compressed dataset $\datasetY_r$ containing the modes of the dataset is constructed through successive iterations using the update rule:
\begin{align}
  \pointy^{t}_{r, i} \leftarrow \frac{ \sumoverkernel{j=1}{|\datasetY|}{\pointy^{t-1}_{r,i}}{\pointy^{0}_j} \pointy^{0}_j}{\sumoverkernel{j=1}{|\datasetY|}{\pointy^{t-1}_i}{\pointy^{0}_j}},
  \label{eq:gms}
\end{align}
where, the indices $i$ and $j$ iterate over each point in $\datasetY$ and the index $t \in \{ 0, \ldots, T-1 \}$
is used to indicate a GMS iteration. At $t = 0$, the compressed dataset is initialized with the
original dataset $\datasetY^0_r = \datasetY^0$. \blue{The convergence criteria for GMS is based on the relative change in
the points between successive iterations or the maximum number of iterations
$T$~\citep{carreira-perpinan_fast_2006}.} At the final iteration, the dataset
contains many overlapping points (indicating modes) and is filtered based on
Euclidean distances to obtain the final output $\datasetY_r$.

\begin{figure}
  \centering
  \subfloat[\label{sfig:ms-low-var}Image A]{\includegraphics[width=0.25\textwidth]{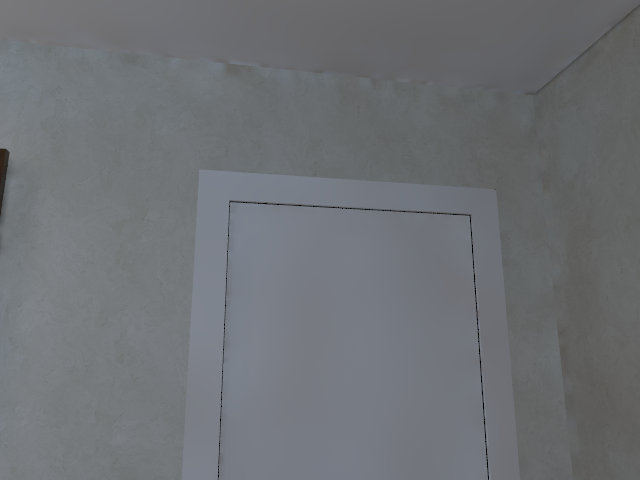}}%
  \subfloat[\label{sfig:ms-high-var}Image B]{\includegraphics[width=0.25\textwidth]{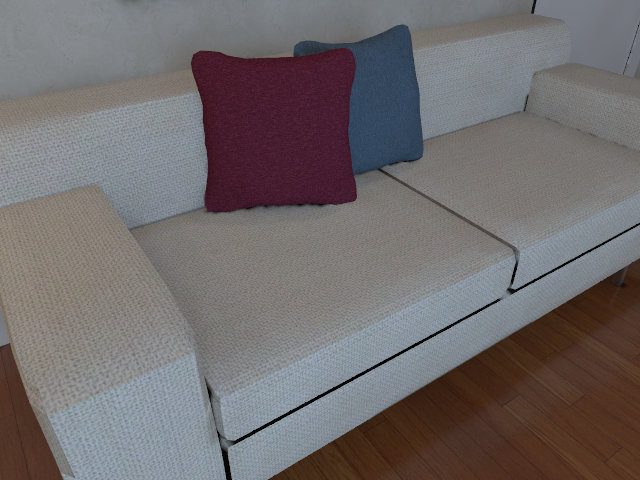}}\\
  \subfloat[\label{sfig:ms-comp-var}Estimated Number of Components]{%
    \ifthenelse{\equal{\arxivmode}{true}}%
    {\includegraphics[]{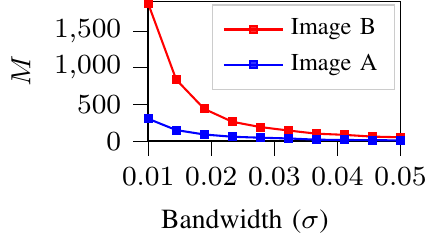}}%
    {\input{./data_analysis/figures/mean_shift_approach_fig.tex}}%
  }%
  \subfloat[\label{sfig:ms-recon-var}Recon. Err. Variation]{%
    \ifthenelse{\equal{\arxivmode}{true}}%
    {\includegraphics[]{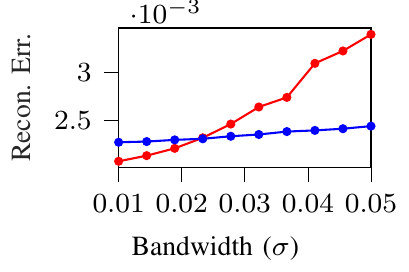}}%
    {\input{./data_analysis/figures/recon_err_fig.tex}}%
  }\\
  \subfloat[\label{sfig:ms-recon-bad}Reconstruction, $\sigma = 0.05$]
  {\ifthenelse{\equal{\arxivmode}{true}}%
  {\includegraphics[width=0.25\textwidth,trim={240 170 270 110},clip]{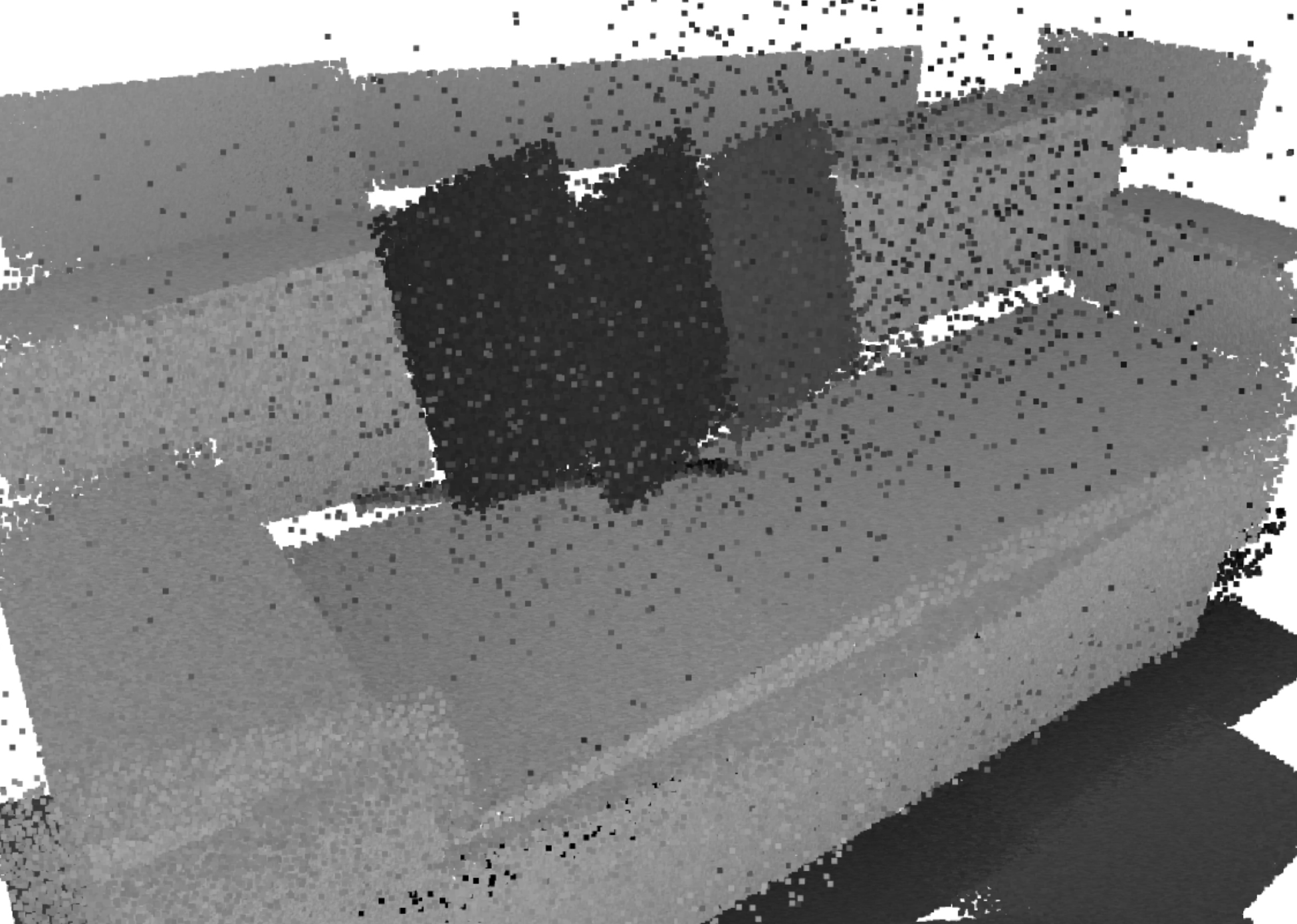}}%
  {\includegraphics[width=0.25\textwidth,trim={240 170 270 110},clip]{./images/couch_0.05.eps}}}%
  \subfloat[\label{sfig:ms-recon-good}Reconstruction, $\sigma = 0.01$]
  {\ifthenelse{\equal{\arxivmode}{true}}%
  {\includegraphics[width=0.25\textwidth,trim={240 170 270 110},clip]{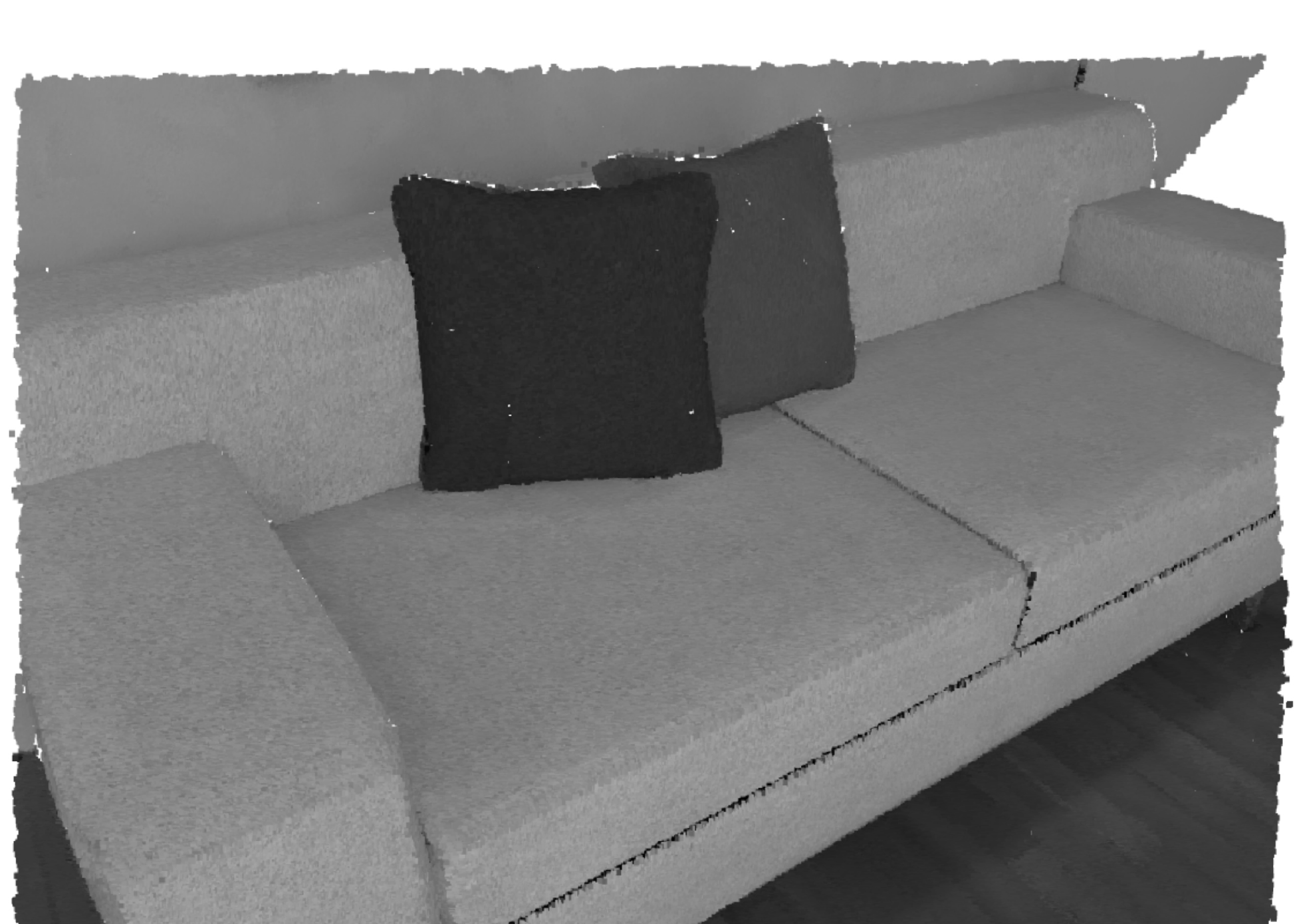}}%
  {\includegraphics[width=0.25\textwidth,trim={240 170 270 110},clip]{./images/couch_0.01.eps}}}
  \caption{\label{fig:ms-output-comparison}
A study on how the PRI (\cref{eq:model}) component in the SOGMM
system~(\cref{fig:overview}) adapts the model size according to the
scene complexity. A simple scene consisting of
\protect\subref{sfig:ms-low-var} homogeneous, white walls requires
fewer components than a~\protect\subref{sfig:ms-high-var} complex
scene consisting of discrete, structured
objects.~\protect\subref{sfig:ms-comp-var} plots the number of
components required to represent each of the two scenes for a given
bandwidth parameter, $\sigma$.~\protect\subref{sfig:ms-recon-var}
plots the mean reconstruction error variation with $\sigma$.~\protect\subref{sfig:ms-recon-bad}
and~\protect\subref{sfig:ms-recon-good} show the reconstruction result for the extrema bandwidths
in~\protect\subref{sfig:ms-recon-var}. Note how the SOGMM formulation
selects more components to represent the complex scene for a given
bandwidth value. Further, the reconstruction error varies monotonically with $\sigma$.}

\vspace{-0.5cm}
\end{figure}

\textbf{Mean Shift on Image Pair.} For the SOGMM system, \blue{we set} the value of $M$
for a given image pair $\imagepair$ \blue{to be} the number of unique points in the
reduced dataset, i.e., $M = | \datasetY_r |$. In contrast to applying the mean
shift algorithm in $\realspace{4}$ using the point cloud $\pointcloud$, we
create the dataset $\datasetY$ from~\cref{eq:pri} in $\realspace{2}$ space using
depth and grayscale values from the image pair $\imagepair$. Each point
$\pointy$ in the dataset is a tuple $(d_i, g_i)$, such that $d_i \in \dimage$,
$g_i \in \gimage$, and $i$ is an index for a pixel coordinate in the images.
Thus, the size of the dataset $\datasetY$ is $hw$ (equal to that of
$\pointcloud$) and $\dimension=2$.

The time complexity of the GMS algorithm increases quadratically with the number of
points, linearly with the number of iterations, and linearly with the dimension
of the data, $\bigoh(TD|\datasetY|^2)$~\citep{jang_meanshift_2021}. The number
of iterations for convergence $T$ depends on the bandwidth value $\sigma$. In general,
lower values of $\sigma$ require larger values of $T$. For faster
convergence,~\citet{carreira-perpinan_fast_2006} propose a modification to the
update rule in~\cref{eq:gms} called the Gaussian Blurring Mean Shift
(GBMS) algorithm and an early stopping criteria to obtain similar results.
Instead of
using the iterate $\pointy^{0}_j$ for every iteration in~\cref{eq:gms}, GBMS
uses the result from the previous iteration $\pointy^{t-1}_j$:
\begin{align}
  \pointy^{t}_{r, i} \leftarrow \frac{ \sumoverkernel{j=1}{|\datasetY|}{\pointy^{t-1}_{r,i}}{\pointy^{t-1}_j} \pointy^{t-1}_j}{\sumoverkernel{j=1}{|\datasetY|}{\pointy^{t-1}_i}{\pointy^{t-1}_j}}.
  \label{eq:gbms}
\end{align}

Further,~\citet{comaniciu_mean_2002} show that utilizing a flat kernel as
opposed to a Gaussian kernel also produces reasonable accuracy in applications
like image segmentation while saving substantial computation due to the finite
support of the flat kernel. As opposed to the Gaussian kernel in~\cref{eq:gaussiankernel},
the flat kernel is given by:
\begin{align}
  \kernel{\pointy - \pointy_i} = \begin{cases} 1 & \text{if } \| \pointy - \pointy_i \| \leq \sigma \\ 0 &\text{if } \| \pointy - \pointy_i \| > \sigma \end{cases}.
  \label{eq:flatkernel}
\end{align}
In this work, we utilize both of these
approximations to make PRI tractable for dense image pair data.

\Cref{fig:ms-output-comparison} illustrates a result of applying the
method on image pairs corresponding to a scene with low variation in depth and
intensity (\cref{sfig:ms-low-var}) and a scene with high variation in depth and
intensity (\cref{sfig:ms-high-var}). For increasing values of the bandwidth
parameter $\sigma$ used by the kernel in~\cref{eq:gms}, we observe a monotonic
decrease in the estimated number of components (\cref{sfig:ms-comp-var}) and
increase in the mean reconstruction error (\cref{sfig:ms-recon-var,sfig:ms-recon-bad,sfig:ms-recon-good}).
Further, for the scene in~\cref{sfig:ms-high-var}, the estimated value of $M$ is
higher for all bandwidth values compared to the scene in~\cref{sfig:ms-low-var}.
This behavior is desired for the SOGMM system as the value of $M$ in the model
given by~\cref{eq:model} must adapt automatically according to the scene
complexity.  Thus, using this system, adaptive complexity in GMM-based point
cloud modeling can be achieved by only specifying the bandwidth parameter
$\sigma$. The choice of bandwidth parameter can be based on the amount of
computation available and the level of fidelity in the GMM model required by the
application.
\section{EVALUATION}\label{sec:evaluation}
\ifthenelse{\equal{\arxivmode}{true}}%
{
\begin{figure*}
  \centering
  \subfloat[\textit{Wall} (original)\label{sfig:stonewall}]{\includegraphics[width=0.33\textwidth,trim={400 200 440 140},clip]{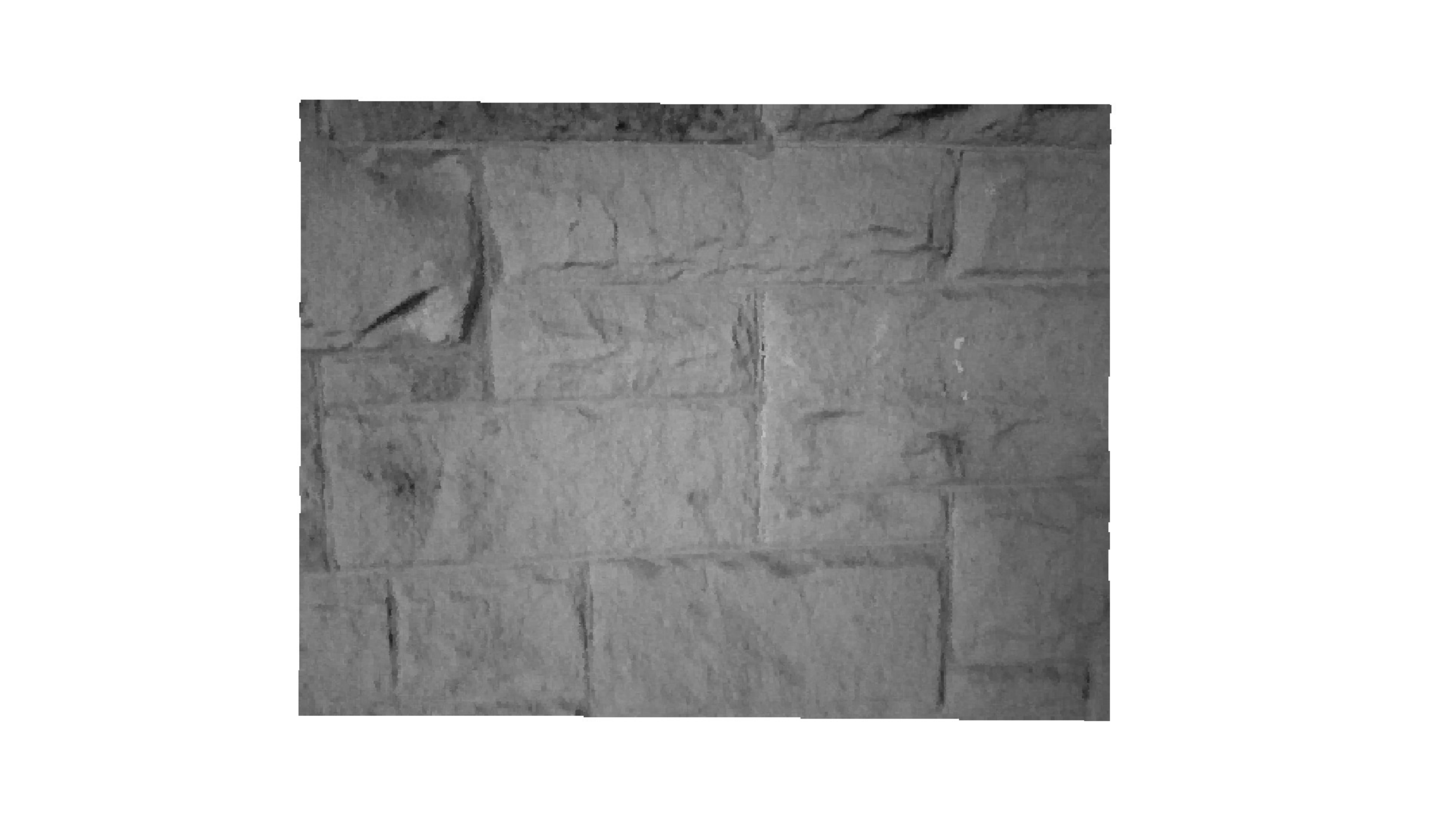}}%
  \subfloat[\textit{Copier} (original)\label{sfig:copyroom}]{\includegraphics[width=0.33\textwidth,trim={400 200 440 140},clip]{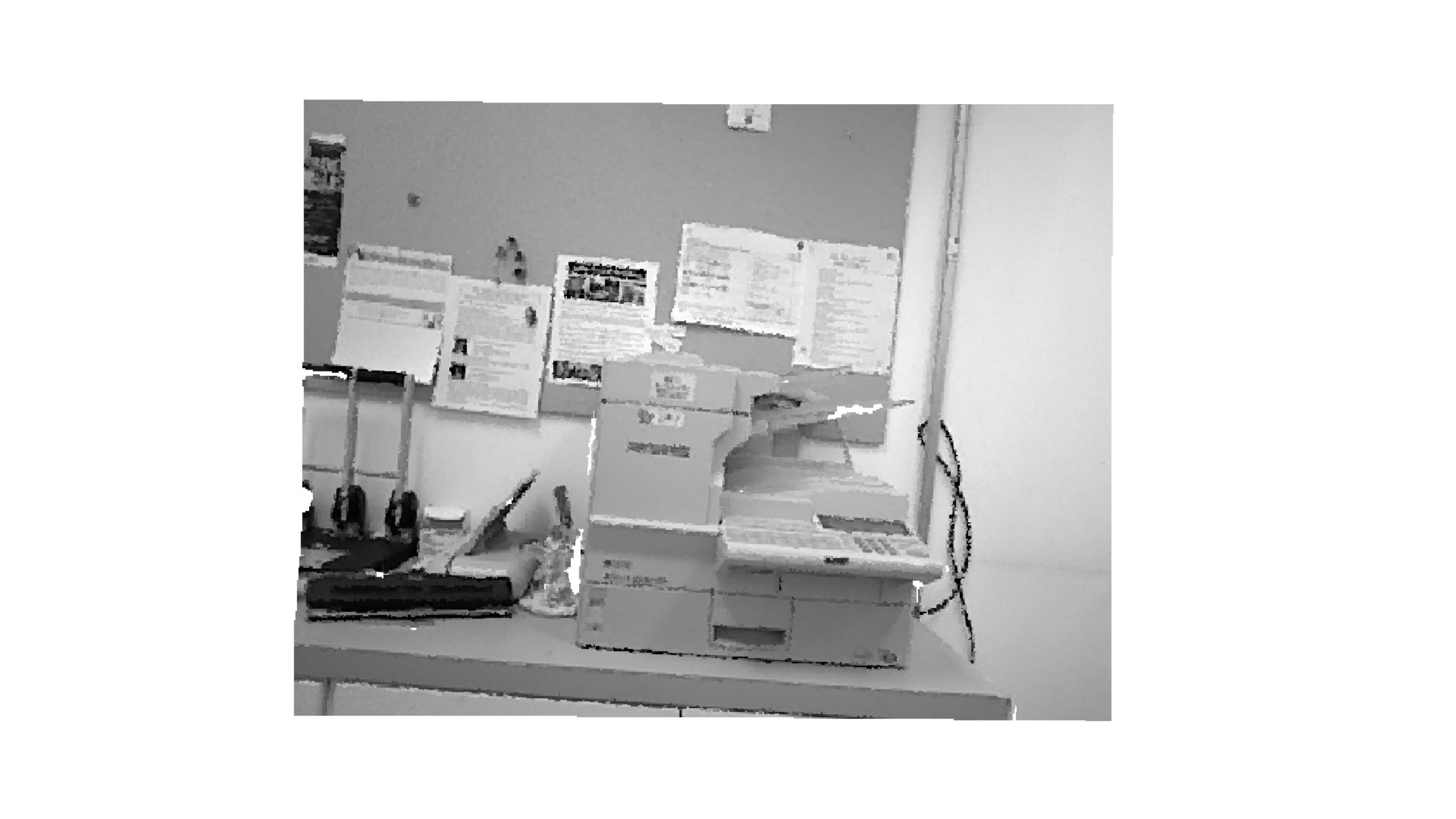}}%
  \subfloat[\textit{Plant} (original)\label{sfig:lounge}]{\includegraphics[width=0.33\textwidth,trim={400 200 440 140},clip]{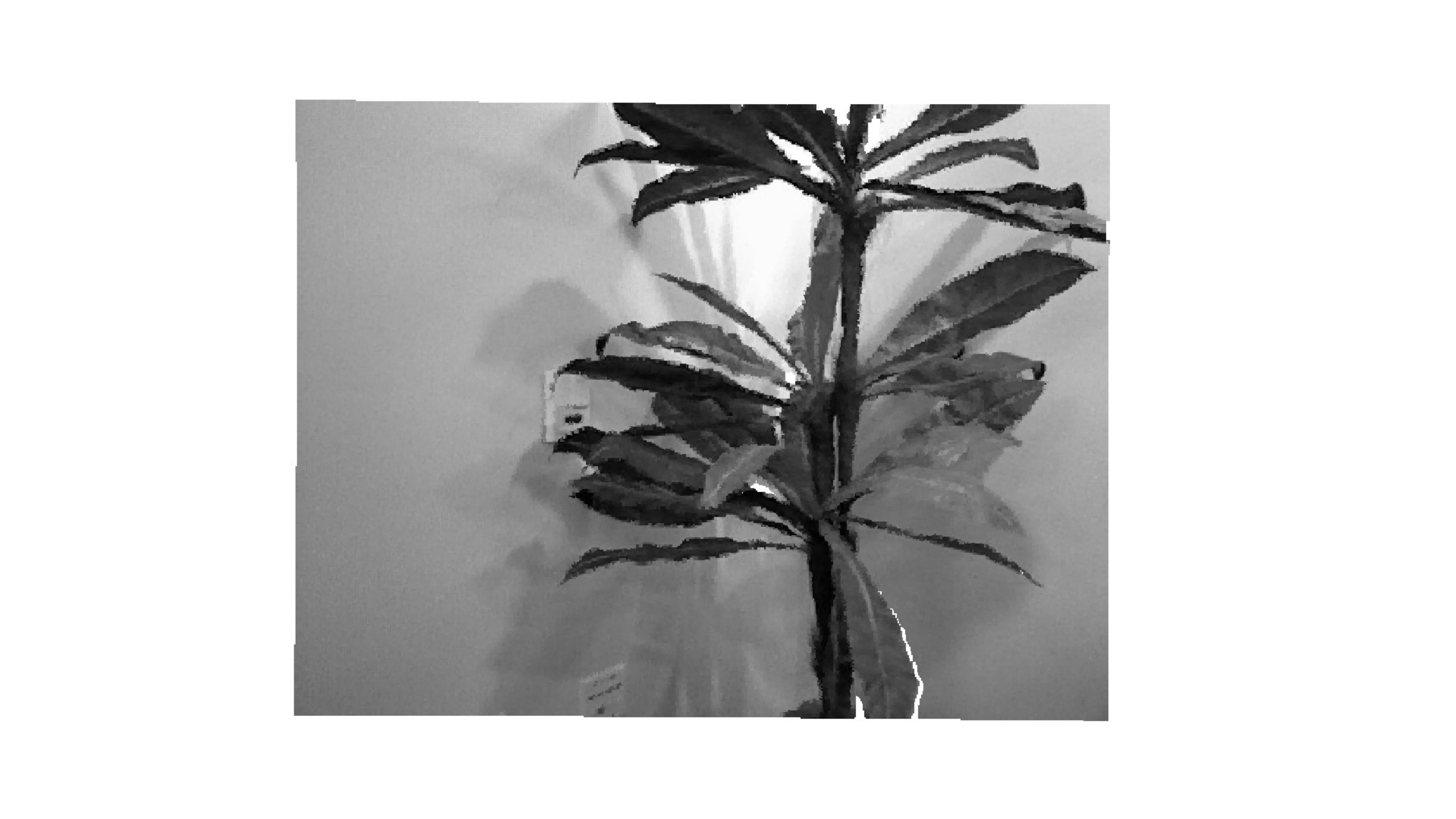}}\\
  \subfloat[\textit{Wall} (reconstruction), OctoMap $\SI{0.02}{\meter}$\label{sfig:stonewall_ot}]{\includegraphics[width=0.33\textwidth,trim={400 200 440 140},clip]{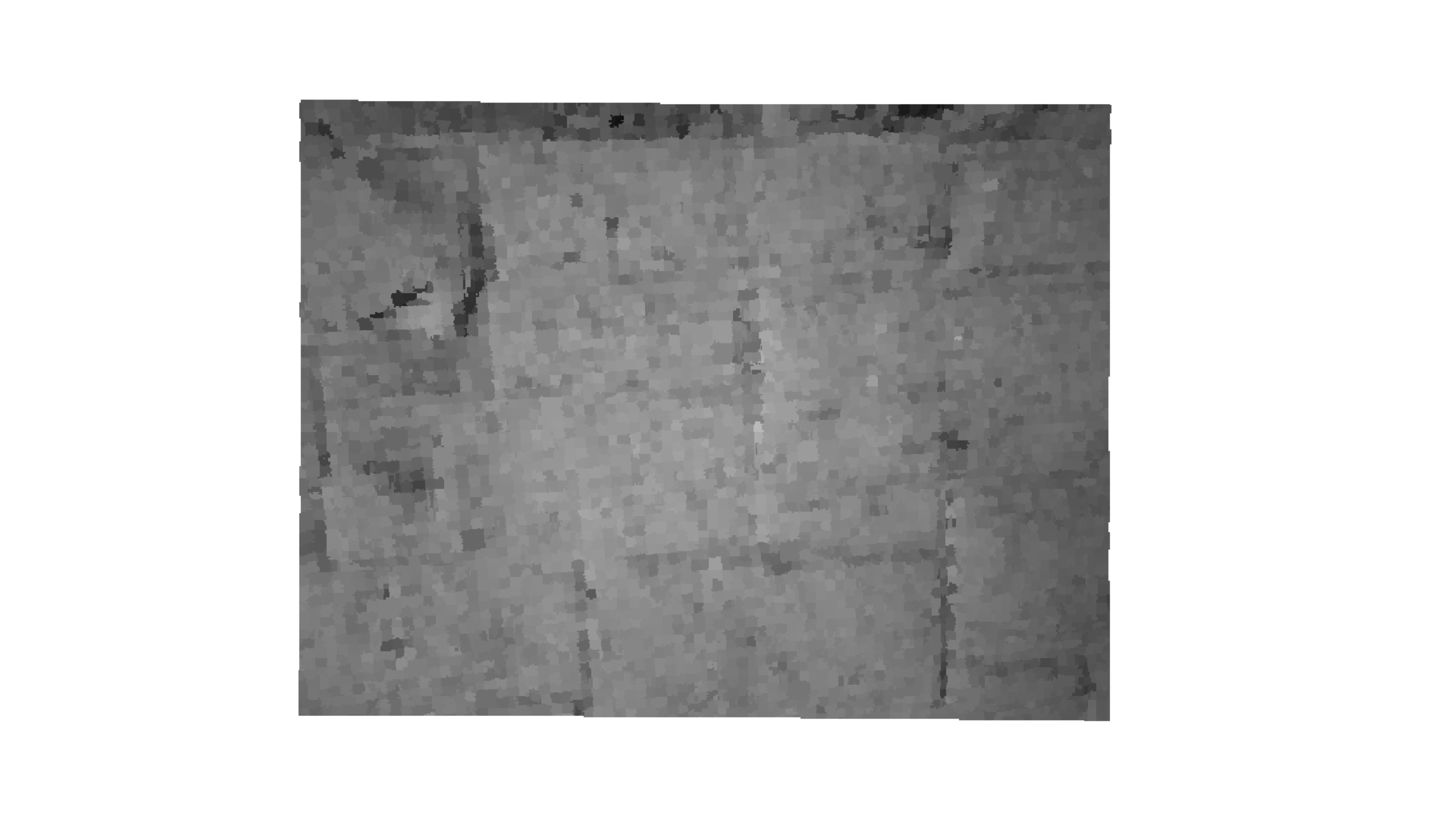}}%
  \subfloat[\textit{Copier} (reconstruction), OctoMap $\SI{0.02}{\meter}$\label{sfig:copyroom_ot}]{\includegraphics[width=0.33\textwidth,trim={400 200 440 140},clip]{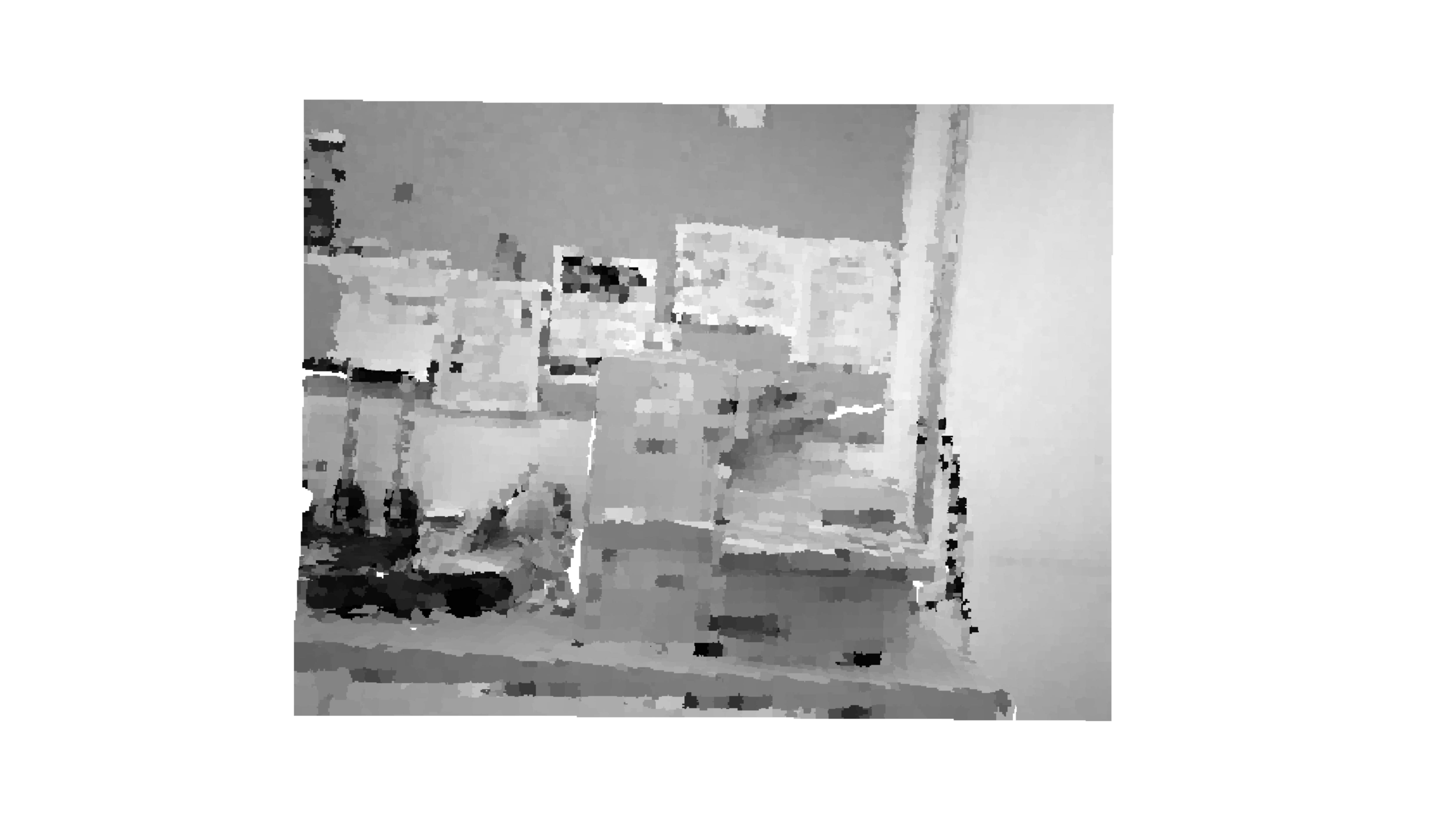}}%
  \subfloat[\textit{Plant} (reconstruction), OctoMap $\SI{0.02}{\meter}$\label{sfig:lounge_ot}]{\includegraphics[width=0.33\textwidth,trim={400 200 440 140},clip]{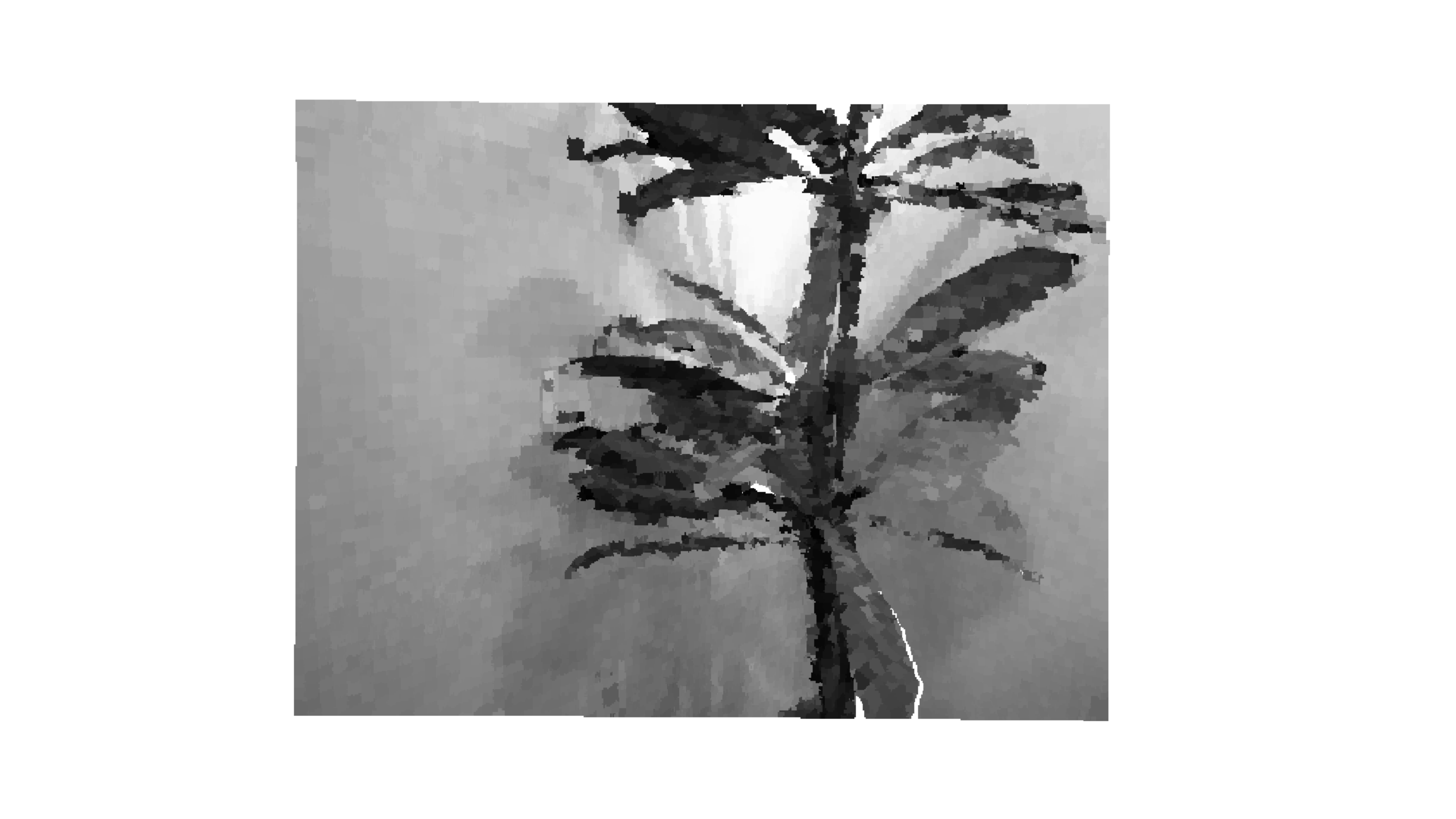}}\\
  \subfloat[\textit{Wall} (reconstruction), NDTMap $\SI{0.02}{\meter}$\label{sfig:stonewall_ndt}]{\includegraphics[width=0.33\textwidth,trim={400 200 440 140},clip]{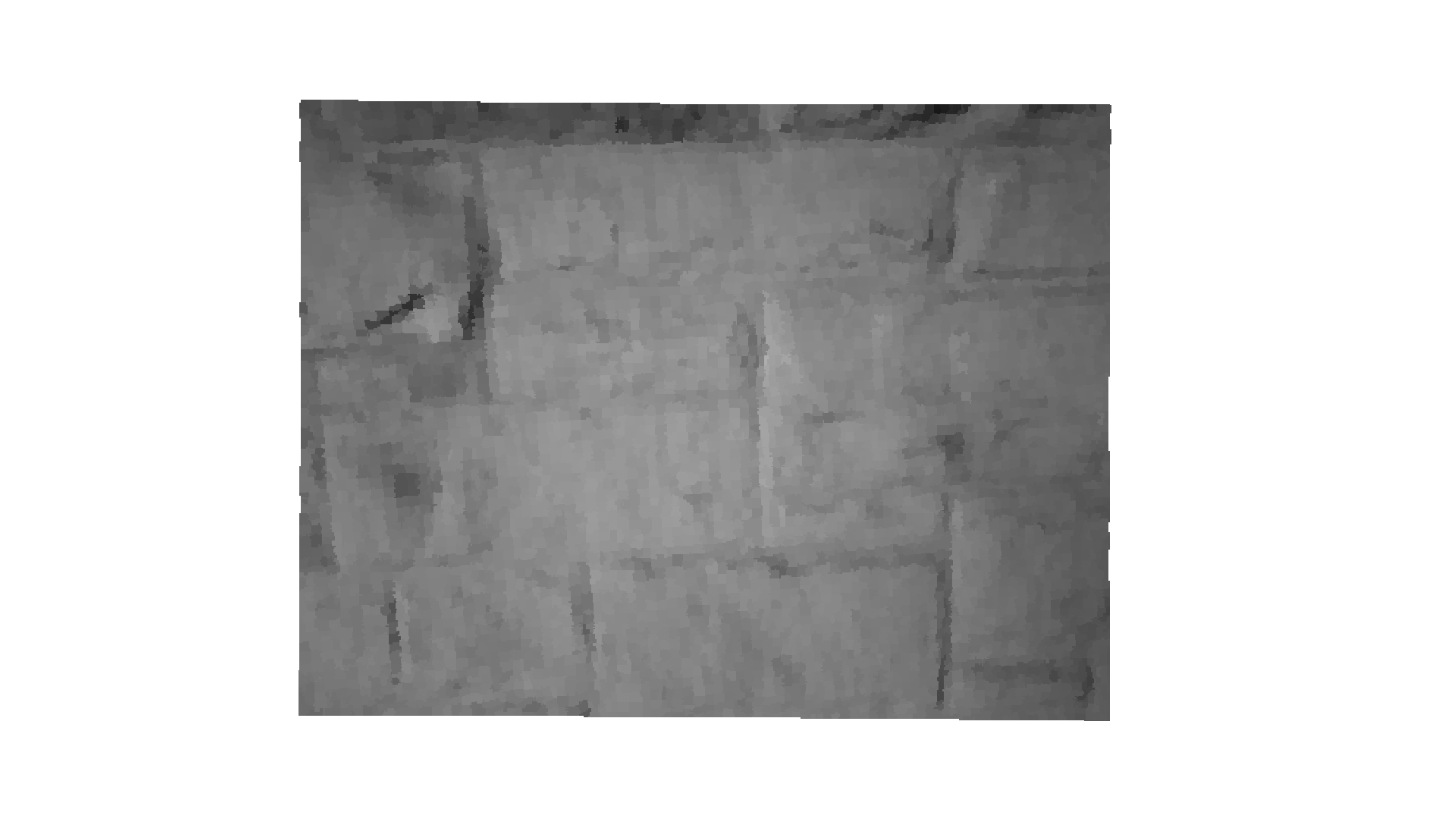}}%
  \subfloat[\textit{Copier} (reconstruction), NDTMap $\SI{0.02}{\meter}$\label{sfig:copyroom_ndt}]{\includegraphics[width=0.33\textwidth,trim={400 200 440 140},clip]{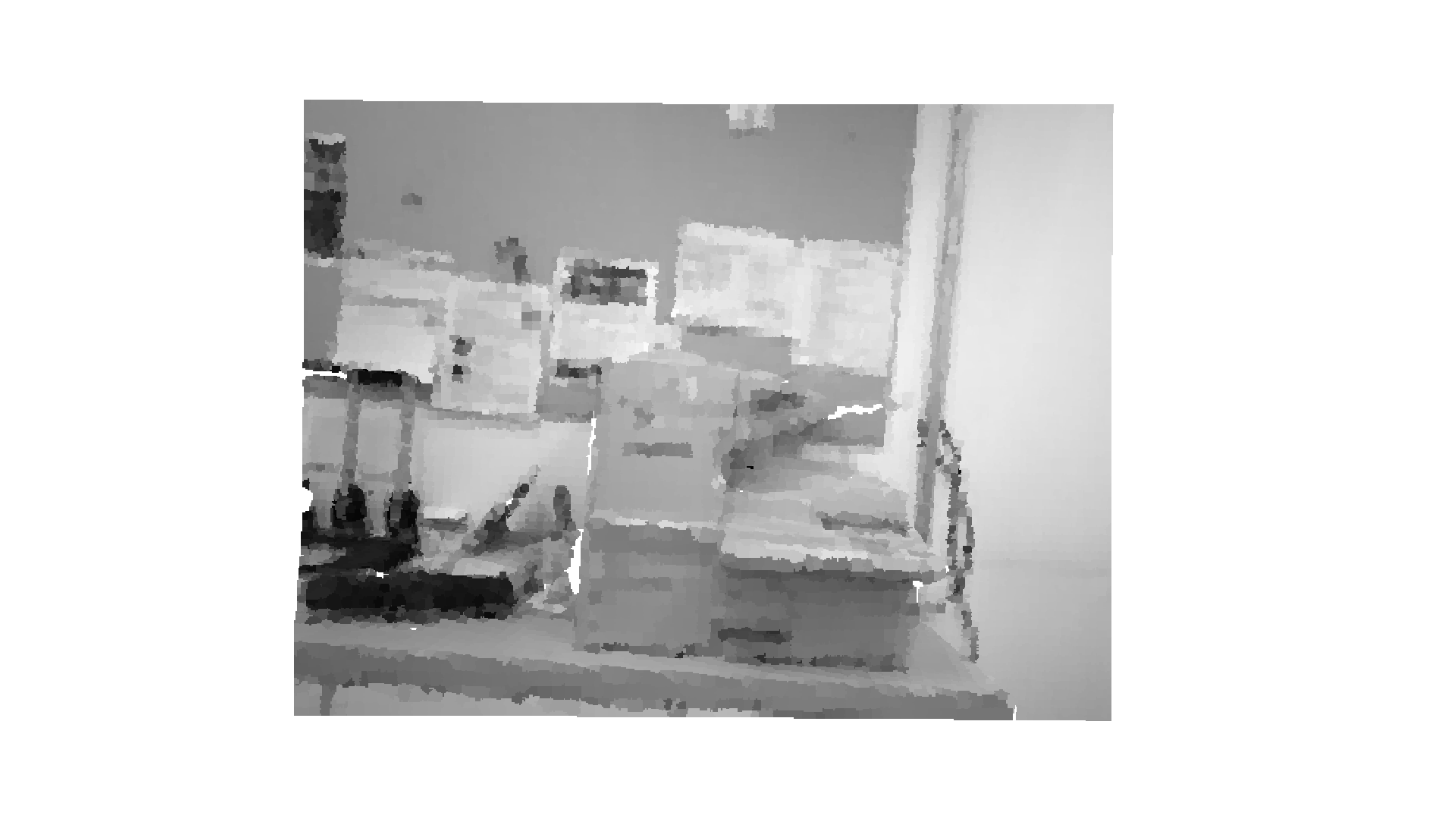}}%
  \subfloat[\textit{Plant} (reconstruction), NDTMap $\SI{0.02}{\meter}$\label{sfig:lounge_ndt}]{\includegraphics[width=0.33\textwidth,trim={400 200 440 140},clip]{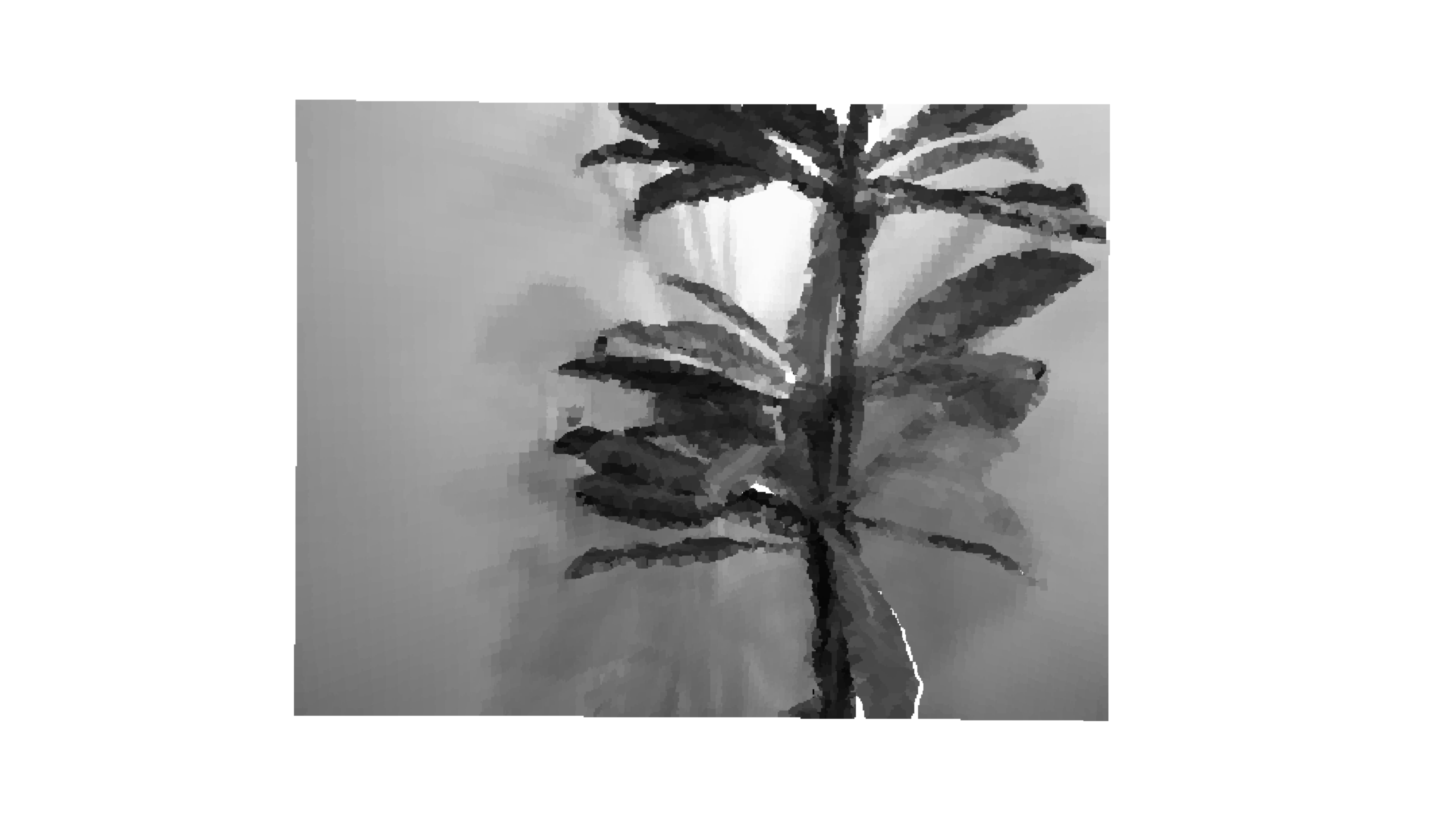}}\\
  \subfloat[\textit{Wall} (reconstruction), $\sigma = 0.01$\label{sfig:stonewall_recon}]{\includegraphics[width=0.33\textwidth,trim={400 200 440 140},clip]{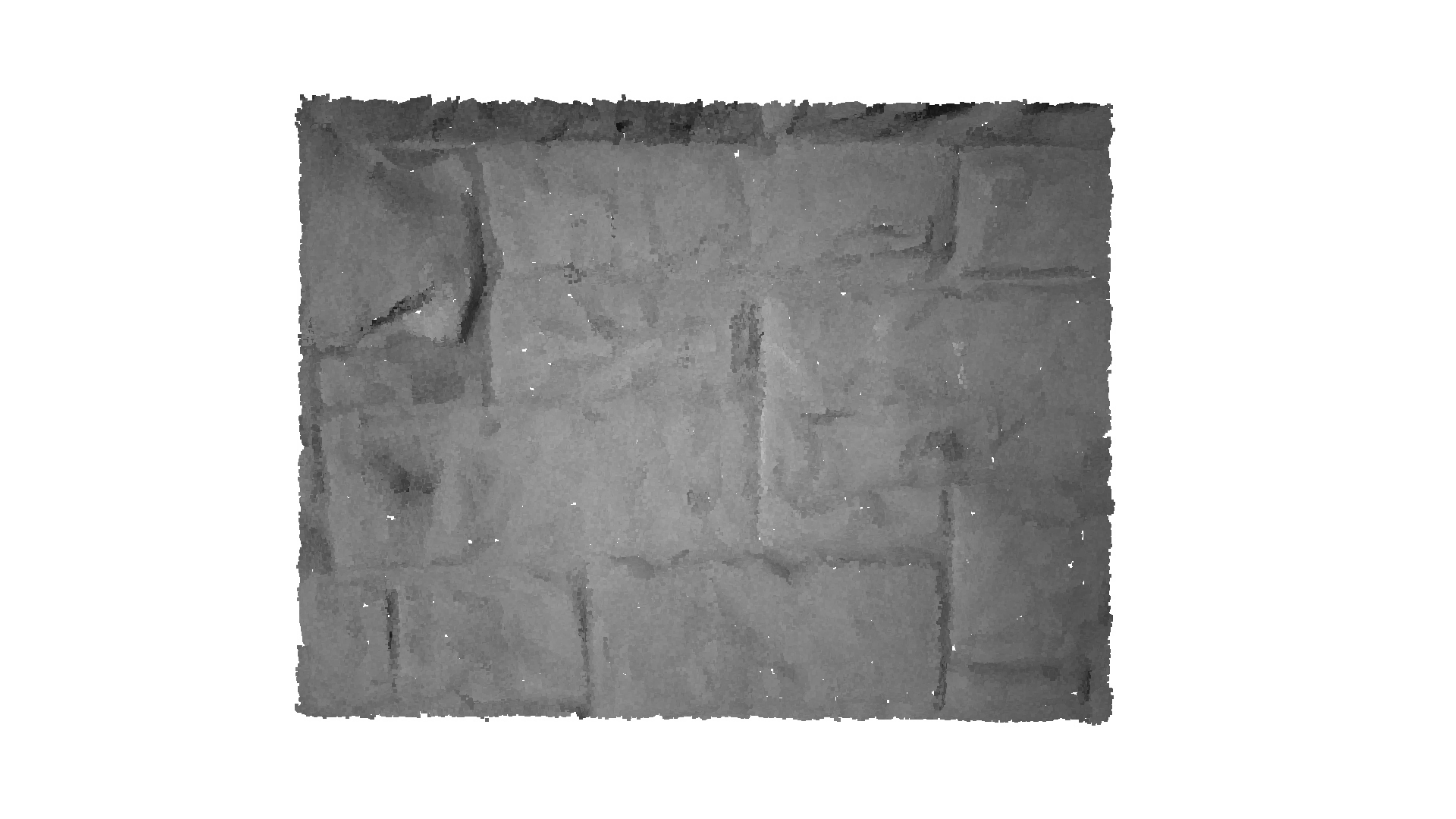}}%
  \subfloat[\textit{Copier} (reconstruction), $\sigma = 0.01$\label{sfig:copyroom_recon}]{\includegraphics[width=0.33\textwidth,trim={400 200 440 140},clip]{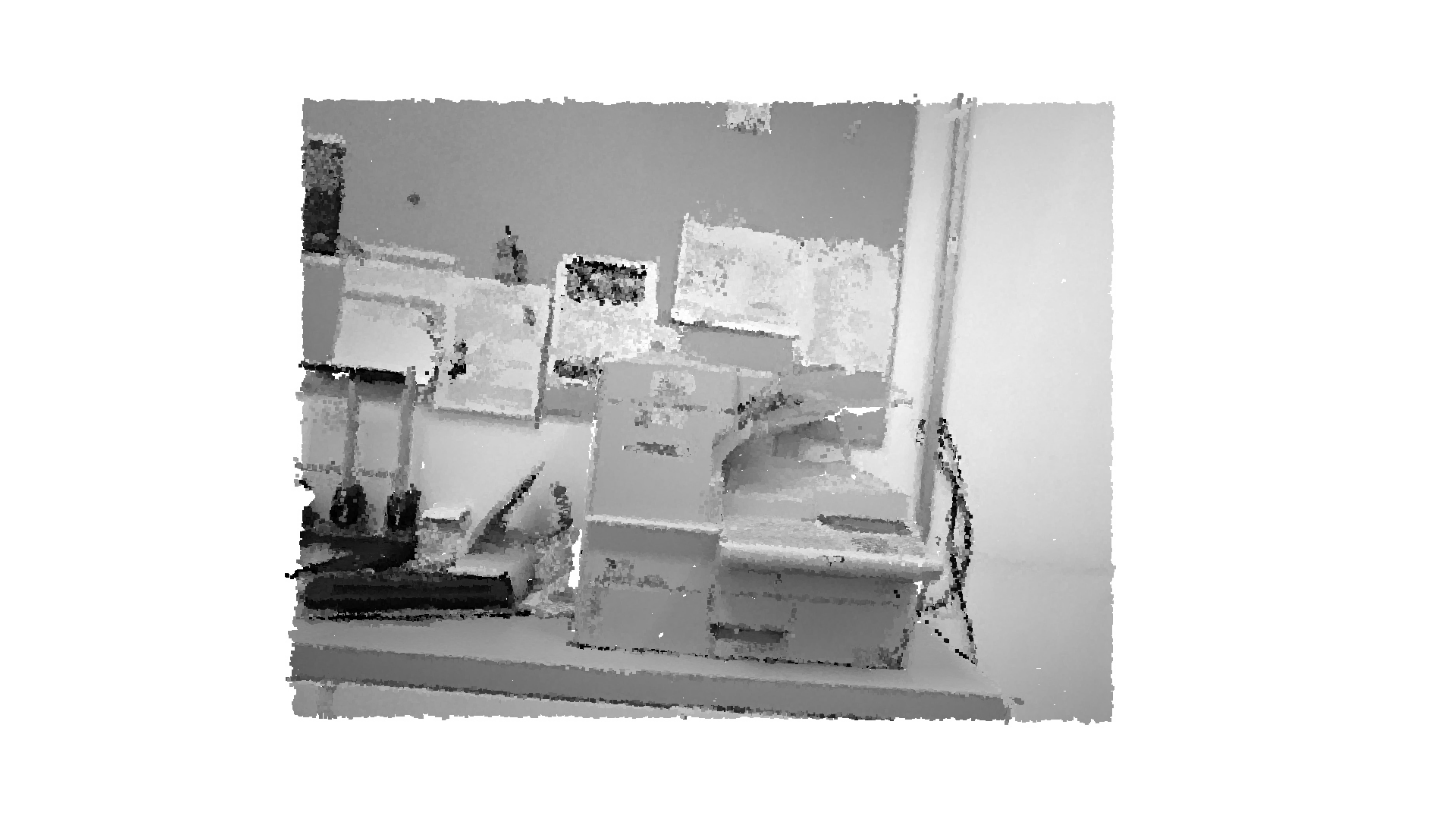}}%
  \subfloat[\textit{Plant} (reconstruction), $\sigma = 0.01$\label{sfig:lounge_recon}]{\includegraphics[width=0.33\textwidth,trim={400 200 440 140},clip]{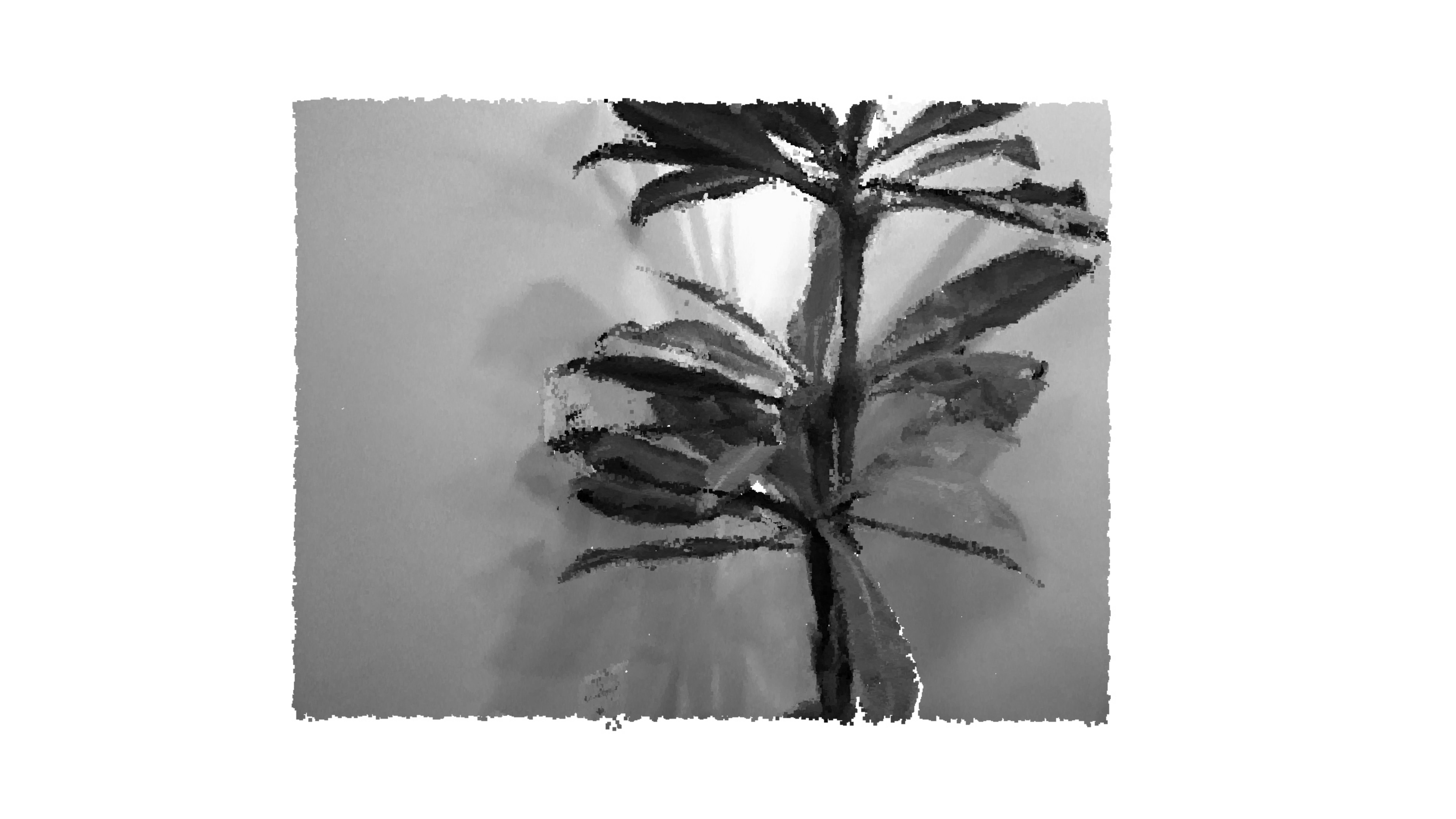}}
  \caption{\label{fig:single-view-qual}Resampled output from SOGMMs created for
      three point clouds with different levels of complexity. The point clouds are
      taken from real-world datasets~\citep{zhou_dense_2013}. The OctoMap method
      results in a pixelated output. NDTMap allows a smoother output at a cost of
      high memory usage. The SOGMM method
      adapts the complexity of the mixture model without changing parameters across
      different scenes ($\sigma = 0.01$ for all the cases). A supplementary video may
      be found at \url{https://youtu.be/v0DfhK1lyno}.
  }
\end{figure*}
}%
{
\begin{figure*}
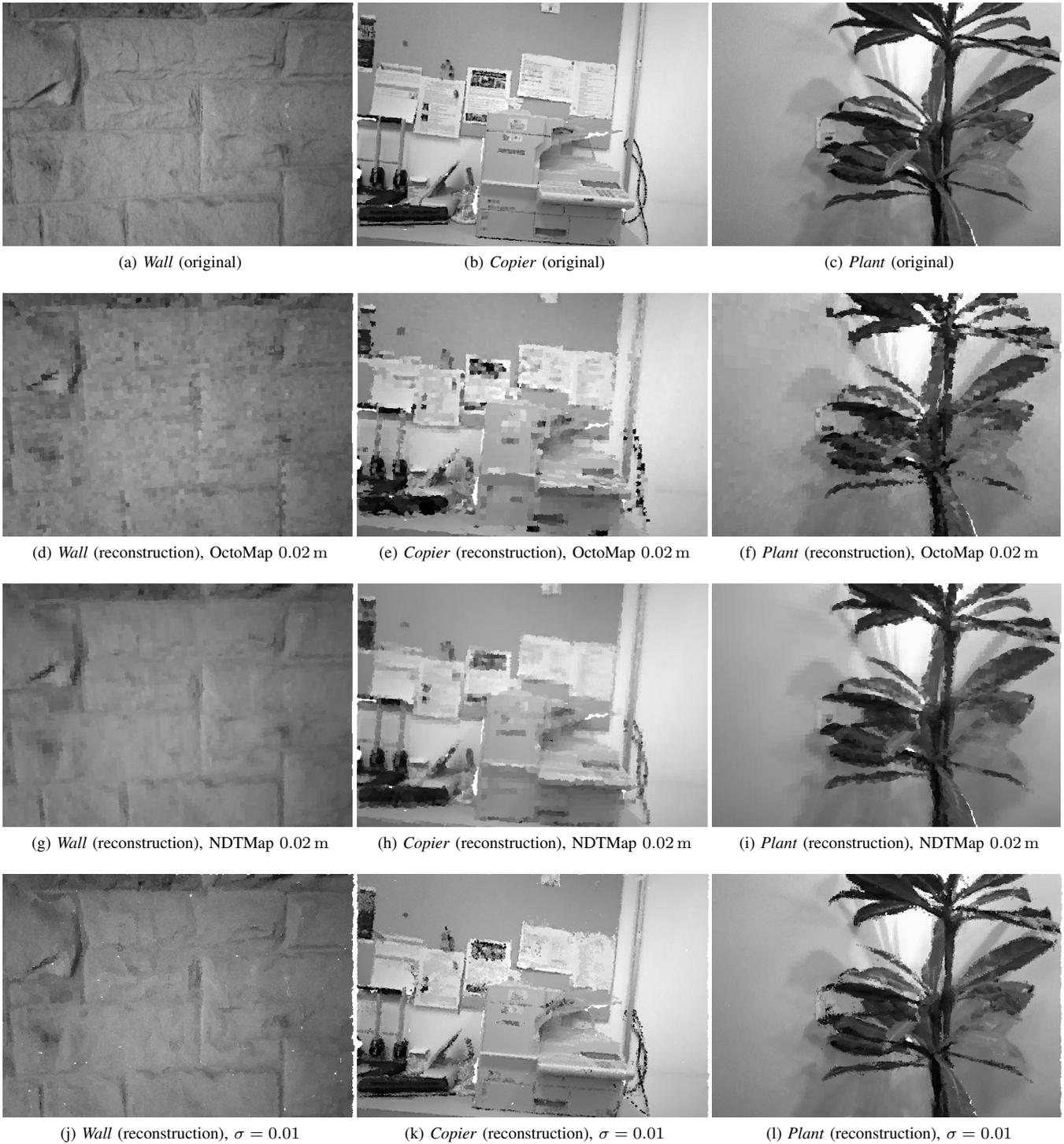

  \centering
  \subfloat[\textit{Wall} (original)\label{sfig:stonewall}]{\includegraphics[width=0.33\textwidth,trim={400 200 440 140},clip]{images/wall_gt.eps}}%
  \subfloat[\textit{Copier} (original)\label{sfig:copyroom}]{\includegraphics[width=0.33\textwidth,trim={400 200 440 140},clip]{images/copier_gt.eps}}%
  \subfloat[\textit{Plant} (original)\label{sfig:lounge}]{\includegraphics[width=0.33\textwidth,trim={400 200 440 140},clip]{images/plant_gt.eps}}\\
  \subfloat[\textit{Wall} (reconstruction), OctoMap $\SI{0.02}{\meter}$\label{sfig:stonewall_ot}]{\includegraphics[width=0.33\textwidth,trim={400 200 440 140},clip]{images/wall_recon_ot.eps}}%
  \subfloat[\textit{Copier} (reconstruction), OctoMap $\SI{0.02}{\meter}$\label{sfig:copyroom_ot}]{\includegraphics[width=0.33\textwidth,trim={400 200 440 140},clip]{images/copier_recon_ot.eps}}%
  \subfloat[\textit{Plant} (reconstruction), OctoMap $\SI{0.02}{\meter}$\label{sfig:lounge_ot}]{\includegraphics[width=0.33\textwidth,trim={400 200 440 140},clip]{images/plant_recon_ot.eps}}\\
  \subfloat[\textit{Wall} (reconstruction), NDTMap $\SI{0.02}{\meter}$\label{sfig:stonewall_ndt}]{\includegraphics[width=0.33\textwidth,trim={400 200 440 140},clip]{images/wall_recon_ndt.eps}}%
  \subfloat[\textit{Copier} (reconstruction), NDTMap $\SI{0.02}{\meter}$\label{sfig:copyroom_ndt}]{\includegraphics[width=0.33\textwidth,trim={400 200 440 140},clip]{images/copier_recon_ndt.eps}}%
  \subfloat[\textit{Plant} (reconstruction), NDTMap $\SI{0.02}{\meter}$\label{sfig:lounge_ndt}]{\includegraphics[width=0.33\textwidth,trim={400 200 440 140},clip]{images/plant_recon_ndt.eps}}\\
  \subfloat[\textit{Wall} (reconstruction), $\sigma = 0.01$\label{sfig:stonewall_recon}]{\includegraphics[width=0.33\textwidth,trim={400 200 440 140},clip]{images/wall_fit.eps}}%
  \subfloat[\textit{Copier} (reconstruction), $\sigma = 0.01$\label{sfig:copyroom_recon}]{\includegraphics[width=0.33\textwidth,trim={400 200 440 140},clip]{images/copier_fit.eps}}%
  \subfloat[\textit{Plant} (reconstruction), $\sigma = 0.01$\label{sfig:lounge_recon}]{\includegraphics[width=0.33\textwidth,trim={400 200 440 140},clip]{images/plant_fit.eps}}
  \caption{\label{fig:single-view-qual}Resampled output from SOGMMs created for
      three point clouds with different levels of complexity. The point clouds are
      taken from real-world datasets~\citep{zhou_dense_2013}. The OctoMap method
      results in a pixelated output. NDTMap allows a smoother output at a cost of
      high memory usage. The SOGMM method
      adapts the complexity of the mixture model without changing parameters across
      different scenes ($\sigma = 0.01$ for all the cases). A supplementary video may
      be found at \url{https://youtu.be/v0DfhK1lyno}.
  }
\end{figure*}
}
In this section, the performance of the SOGMM method is evaluated on real-world
point cloud data provided by~\citet{zhou_dense_2013}:~\texttt{stonewall}
(\cref{sfig:stonewall}),~\texttt{copyroom} (\cref{sfig:copyroom}),
and~\texttt{lounge} (\cref{sfig:lounge}). Because the proposed
methodology addresses efficient and compact perceptual modeling within
the robotics context, it is compared against four open source mapping
baselines: (1) OctoMap (OM)~\citep{hornung_octomap_2013}, (2) NDTMap (NDT)~\citep{magnusson_scan_2007},
\blue{(3) GPOctoMap (GPOM)~\citep{wang_fast_2016}}, and
(4) GMM with fixed number of components (FC)~\citep{tabib_autonomous_2021}. The
voxel resolutions for the first two approaches are set to \SI{0.02}{\meter} and
\SI{0.05}{\meter}, which have been demonstrated to be adequate for scene
representation~\cite{yan_online_2021}. \blue{A resolution of \SI{0.02}{\meter}
is used for GPOctomap. The default maximum variance parameter is lowered to $0.001$
from the default $0.02$ for better reconstruction quality.} For this section only, the
following shorthand is introduced: an OM with \SI{0.05}{\meter} and
\SI{0.02}{\meter} leaf sizes will be referred to as \om{0.05} and
\om{0.02}, respectively; an NDTMap with voxel resolutions of
\SI{0.05}{\meter} and \SI{0.02}{\meter} will be referred to as
\ndt{0.05} and \ndt{0.02}, respectively; \blue{a GPOM with \SI{0.02}{\meter} will be referred to as \gpom{0.02}};
the GMM approach with
75, 500, and 2000 components will be referred to as \fc{75},
\fc{500}, and \fc{2000}, respectively; and the SOGMM approach
with bandwidths 0.01, 0.02, and 0.03 will be referred to
as \sogmm{0.01}, \sogmm{0.02}, and \sogmm{0.03}, respectively.

\textbf{Reconstruction from Environment Models.} For OctoMap, the
reconstruction at the minimum leaf size is utilized after modeling
occupied space using the~\texttt{ColorOcTree}
class\footnote{\url{https://github.com/OctoMap/octomap/blob/devel/octomap/include/octomap/ColorOcTree.h}}.
\blue{GPOctoMap does not natively support color in the octree
nodes\footnote{\url{https://github.com/RobustFieldAutonomyLab/la3dm/blob/master/include/gpoctomap/gpoctree_node.h}}, so this support was added for our
analysis\footnote{\url{https://github.com/kshitijgoel007/la3dm/blob/feature/colcon/include/gpoctomap/gpoctree_node.h}}.}
For NDTMap, the reconstruction is obtained by modifying
the~\texttt{NDTCell} class to store the average grayscale value
for the points associated with the
cell\footnote{\url{https://github.com/OrebroUniversity/perception_oru/blob/port-kinetic/ndt_map/include/ndt_map/ndt_cell.h}}.
In both cases, intensity is queried at a 3D
coordinate. For the FC and SOGMM approaches, the 3D reconstruction is
obtained by densely sampling the joint distribution
$\density{\randvarX}{\pointx}$. The grayscale reconstruction is obtained by
regressing the expected value from the conditional distribution $p_{g | \pointx}(g | \pointx)$.
This conditional distribution is obtained as follows. First, for each component $m$
in the model given by~\cref{eq:model}, the mean and covariance can be written as:
\begin{align*}
  \mean_m = \begin{bmatrix}
    \mean_{m, \pointx} \\
    \mean_{m, g}
  \end{bmatrix}%
  \;
  \cov_m = \begin{bmatrix}
    \cov_{m, \pointx \pointx} & \cov_{m, \pointx g}\\
    \cov_{m,  g\pointx} & \cov_{m, gg}
  \end{bmatrix}.
\end{align*}

The conditional distribution $p_{g | \pointx}(g | \pointx)$ can be expressed
using these quantities
as~\citep{sung_gaussian_2004}:
\begin{align}
  p_{g | \pointx}(g | \pointx) = \sum_{m=1}^M w_m(\pointx) \gaussian{g}{\lambda_m(\pointx)}{\nu_m^2},
  \label{eq:conditional}
\end{align}
where
\begin{align*}
  w_m(\pointx) &= \frac{\weight_m \gaussian{\pointx}{\mean_{m, \pointx}}{{\cov_{m, \pointx \pointx}}}}{\sum_{m'=1}^M \weight_{m'} \gaussian{\pointx}{\mean_{m', \pointx}}{\cov_{m', \pointx \pointx}}},\\
  \lambda_m(\pointx) &= \mean_{m, g} + \cov_{m, g \pointx} \cov_{m, \pointx \pointx}^{-1} (\pointx - \mean_{m, \pointx}),
\end{align*}
and
\begin{align*}
  \nu_m^2 &= \cov_{m, gg} - \cov_{m, g \pointx} \cov_{m, \pointx \pointx}^{-1} \cov_{m, \pointx g}.
\end{align*}
Finally, the expected value for~\cref{eq:conditional} is:
\begin{align}
  \lambda(\pointx) = \sum_{m=1}^M w_m(\pointx) \lambda_m(\pointx).
\end{align}

\textbf{Qualitative Evaluation.} \Cref{fig:single-view-qual} provides
a qualitative evaluation of the OctoMap, NDTMap, and
SOGMM.~\Cref{sfig:stonewall,sfig:copyroom,sfig:lounge} provide images
of low, medium, and high complexity scenes,
respectively. \blue{The OctoMap edge effects at the voxel boundaries
prevents} smooth color representation (see~\cref{sfig:stonewall_ot,sfig:copyroom_ot,sfig:lounge_ot}).
The NDTMap representation can be sampled using the Gaussian components in
each voxel, \blue{which} results in a qualitatively better reconstruction
(\cref{sfig:stonewall_ndt,sfig:copyroom_ndt,sfig:lounge_ndt}) compared
to OctoMap at a cost of high memory footprint. The SOGMM
representation is sampled to generate reconstructions shown
in~\cref{sfig:stonewall_recon,sfig:copyroom_recon,sfig:lounge_recon},
respectively. While the bandwidth parameter remains constant across
scenes, the estimated number of components increases with the scene
complexity. \blue{It is evident that the fine details and intensity
are preserved in the edges around the
stones~(\cref{sfig:stonewall_recon}), thin
wires~(\cref{sfig:copyroom_recon}), and in the leaves and
shadows~(\cref{sfig:lounge_recon}).}
The SOGMM framework preserves fine
details by learning the appropriate model complexity from the
underlying sensor data without parameter tuning from the user.

\begin{figure*}
   \centering
  \begin{minipage}{\textwidth}
    \subfloat[PSNR for Grayscale Image Reconstruction (higher is better)\label{sfig:psnr}]{%
      \ifthenelse{\equal{\arxivmode}{true}}%
      {\includegraphics[]{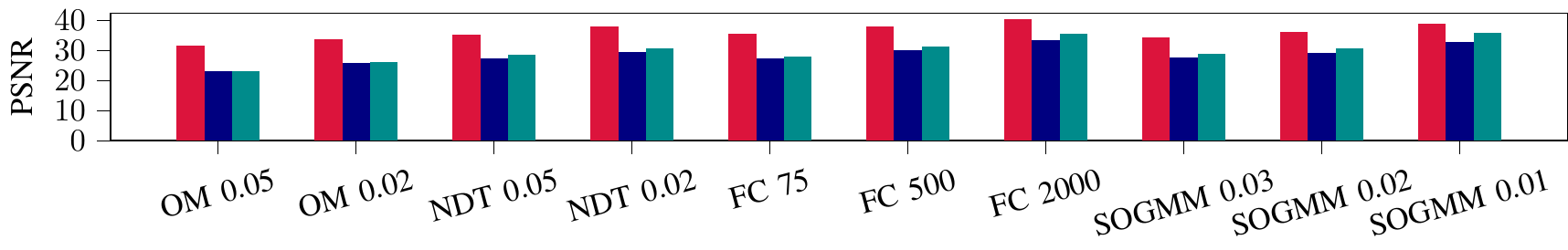}}%
      {\input{data_analysis/figures/single-view-psnr.tex}}%
    }\\
    \subfloat[Mean Reconstruction Error for 3D Reconstruction (lower is better)\label{sfig:recon_err}]{%
      \ifthenelse{\equal{\arxivmode}{true}}%
      {\includegraphics[]{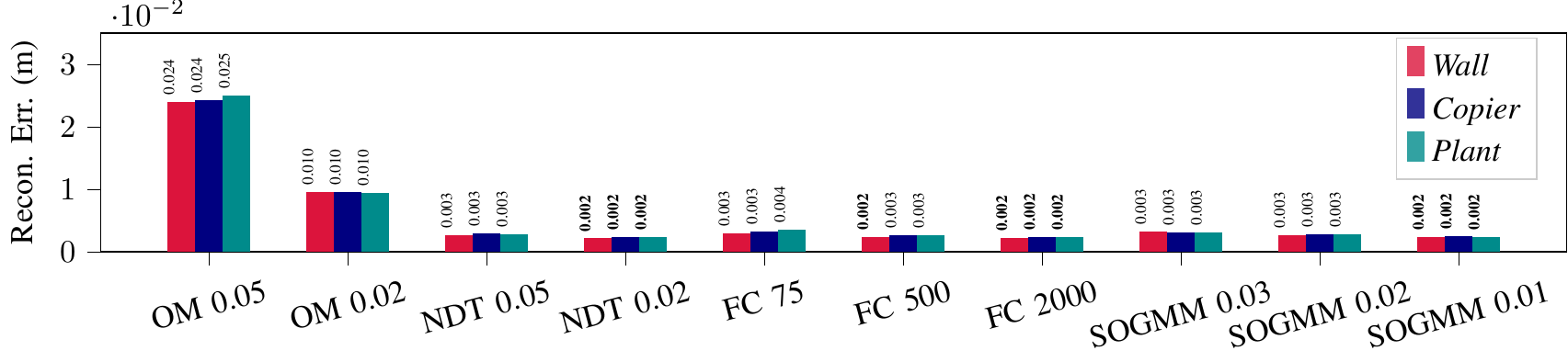}}%
      {\input{data_analysis/figures/single-view-err.tex}}%
    }\\
    \subfloat[Size of the Model (lower is better)\label{sfig:model_size}]{%
      \ifthenelse{\equal{\arxivmode}{true}}%
      {\includegraphics[]{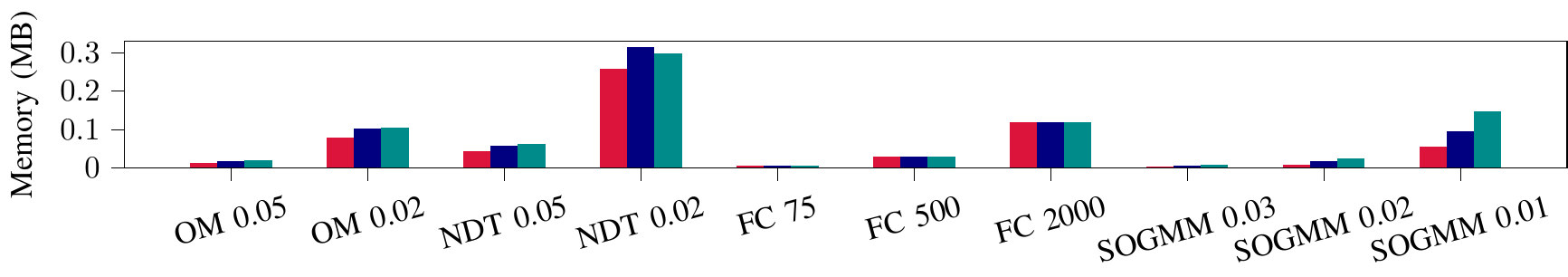}}%
      {\input{data_analysis/figures/single-view-mem.tex}}%
    }\\
  \end{minipage}%
  \caption{\label{fig:single-view-quant} Quantitative evaluation of the SOGMM method for the scenes shown
  in~\cref{fig:single-view-qual}.
  The SOGMM approach enables high accuracy~\protect\subref{sfig:psnr} grayscale
  and~\protect\subref{sfig:recon_err} 3D reconstructions while allowing~\protect\subref{sfig:model_size}
  adaptation in the size of the model without
  changing the bandwidth parameter across scenes.}
\vspace{-0.25cm}
\end{figure*}

\renewcommand{\tabcolsep}{1pt}
\begin{table*}
  \begin{center}
  \begin{tabular}{l*3c|*2c|*4c|*4c|*4c}
    \toprule
	  Dataset & \multicolumn{3}{c}{\om{0.02}} & \multicolumn{2}{c}{\blue{\gpom{0.02}}} &  \multicolumn{4}{c}{\ndt{0.02}} & \multicolumn{4}{c}{\fc{2000}} & \multicolumn{4}{c}{\sogmm{0.01}}\\
    \midrule
     & \shortstack{PSNR\\(dB)} & \shortstack{MRE\\(m)} & \shortstack{Mem.\\(MB)} & \shortstack{PSNR\\(dB)} & \shortstack{MRE\\(m)} & \shortstack{PSNR\\(dB)} & \shortstack{MRE\\(m)} & \shortstack{Mem.\\(MB)} & \# Voxels & \shortstack{PSNR\\(dB)} & \shortstack{MRE\\(m)} & \shortstack{Mem.\\(MB)} & \# Comp. & \shortstack{PSNR\\(dB)} & \shortstack{MRE\\(m)} & \shortstack{Mem.\\(MB)} & \# Comp.\\
    \midrule
    \textit{Wall}   & 33.9 & 0.010& 0.078 & \blue{34.0} & \blue{0.009} & 38.0& \textbf{0.002}&0.26 & 6482 & \textbf{40.5} & \textbf{0.002} & 0.12 & 2000 & 39.1 & \textbf{0.002}& \textbf{0.06} & 933\\
    \textit{Copier} & 25.8 & 0.010& \textbf{0.10} & \blue{25.8} & \blue{0.009} & 30.0& \textbf{0.002}&0.31 & 7856 & \textbf{33.4} & \textbf{0.002} & 0.12 & 2000 & 33.0& \textbf{0.002}& \textbf{0.10} & 1599\\
    \textit{Plant}  & 26.3 & 0.010& \textbf{0.10} & \blue{26.4} & \blue{0.009} & 30.8& \textbf{0.002}&0.30 & 7470 & 35.5 & \blue{\textbf{0.002}} & 0.12 & 2000 & \textbf{36.0}& \textbf{0.002}& 0.15 & 2464\\
    \bottomrule
  \end{tabular}
  \caption{\label{tab:quant-eval} Raw data for the quantitative comparison of the \om{0.02},
  \ndt{0.02}, \fc{2000} and \sogmm{0.01} cases. The SOGMM method
  achieves a balance between the model fidelity and size for diverse scenes
  without changing the single tunable parameter $\sigma$.~\blue{The memory is
    not computed for \gpom{0.02} because the implementation does not
    provide functions to serialize the perceptual model}.}
  \end{center}
  \vspace{-0.5cm}
\end{table*}

\textbf{Quantitative Evaluation.} \blue{The Peak Signal-To-Noise Ratio (PSNR),
  Mean Reconstruction Error (MRE), and (3) Memory Usage
  are used for quantitative evaluation}.  The PSNR quantifies
how accurately intensity is reconstructed from the model by comparing against the
original intensity image. The MRE is computed for the spatial part of
the reconstructed point cloud (i.e., $[x, y, z]^{\top}$) and is given
by the average distance between the reconstructed point cloud and the
ground truth point cloud. The memory usage quantifies the memory
required to store the model in megabytes (MB).  For OctoMap, this
value is the size of the~\texttt{.ot} file that retains the occupancy
and grayscale information. \blue{Note that unlike OctoMap, the
GPOctoMap implementation does not enable serialized storage of the octree. Therefore,
the GPOctoMap approach is compared on the PSNR and MRE metrics\blue{, only}.} Each cell of the NDTMap stores a Gaussian
component and intensity, so the memory usage (assuming floating point
values and $M$ cells) is calculated as $4 \cdot M \cdot (1 + 3 +
6)$ bytes. The FC and SOGMM are composed of $M$ 4D Gaussian
components, so the memory usage is calculated as $4 \cdot M
\cdot (1 + 10 + 4)$ bytes.

\Cref{fig:single-view-quant}~\blue{provides a comparison of
our approach against the OM, NDT, and FC baselines for the \textit{Wall},
\textit{Copier}, and \textit{Plant} scenes shown in~\cref{fig:single-view-qual}}.
The OM baseline exhibits lower PSNR values (\cref{sfig:psnr}) and highest reconstruction
error (\cref{sfig:recon_err}) compared to other approaches. \ndt{0.05} and \ndt{0.02} outperform
\om{0.05} and \om{0.02} in terms of PSNR and reconstruction error but consume larger
amounts of memory (\cref{sfig:model_size}).

While \fc{75}, \fc{500}, and \fc{2000} consume less memory
compared to \ndt{0.02}, \blue{the approach is not adaptive. As a
result,} homogenous scenes will be represented with more fidelity than
required and complex scenes will have insufficient fidelity, which is
undesirable.
In contrast, \blue{the} SOGMM cases \blue{adapt} to use
\blue{fewer} components for \textit{Wall} and progressively higher number\blue{s} of components
for \textit{Copier} and \textit{Plant}. Consequently, we observe similar PSNR and reconstruction
error scores for \sogmm{0.01} and \fc{2000}; however, the former case adapts to
the scene and consumes less memory for the \textit{Wall} and \textit{Copier}
cases.~\Cref{tab:quant-eval} provides raw evaluation data to directly compare \sogmm{0.01}
against \blue{\gpom{0.02}} and the best performing variants of OctoMap, NDT, and FC.
\blue{\gpom{0.02} enables gains in PSNR and MRE compared to \om{0.02} but the accuracy
is lower compared to \blue{the} NDT, FC, and SOGMM approaches.}
\sogmm{0.01} produces a model with accuracy in line with \ndt{0.02},
while consuming significantly less memory. \fc{2000} allows
higher PSNR scores but overestimates the number of components for scenes with
relatively less intensity-depth variation. \sogmm{0.01} on the other hand provides
a variation in the number of components used without changing any parameters across
the three scenes.

\sogmm{0.02} and \sogmm{0.03} show the effect of the bandwidth
parameter $\sigma$.  As expected from~\cref{fig:ms-output-comparison},
the size of the model increases $\sigma$ decreases. The
$\sigma$ parameter can be chosen based on the available computation as
opposed to making an \textit{a priori} guess about the complexity of
the environments being modeled.

\section{CONCLUSION}\label{sec:conclusion}
Computational modeling methods for multi-modal point cloud data lack
the adaptability necessary for generalization across
diverse scenes. This letter detailed a continuous
probabilistic modeling approach that enables adaptation by
estimating the number of components using self-organizing principles
from information-theoretic learning. The quantitative and qualitative
results for the proposed method demonstrate its efficacy on diverse
real-world scenes. Future work includes: (1)
incremental mapping for streaming sensor data, (2) online and anytime
selection of $\sigma$ based on the available compute, (3) a highly
parallelized implementation of the SOGMM system on an embedded GPU,
and (4) enabling cross-modal inference.

{
 \balance
 \footnotesize
 \bibliographystyle{IEEEtranN}
 \bibliography{refs}
}

\end{document}